\documentclass[lettersize,journal]{IEEEtran}
\usepackage{amsmath,amsfonts}
\usepackage{algorithm}
\usepackage{array}
\usepackage[caption=false,font=normalsize,labelfont=sf,textfont=sf]{subfig}
\usepackage{textcomp}
\usepackage{stfloats}
\usepackage{url}
\usepackage{verbatim}
\usepackage{graphicx}
\usepackage{cite}
\usepackage{booktabs}
\usepackage{lscape}
\usepackage{soul, color, xcolor}
\usepackage{multirow}
\usepackage{epstopdf}
\usepackage{algorithm}
\usepackage{algorithmicx}
\usepackage{algpseudocode}
\usepackage{float} 
\usepackage{subfloat}
\usepackage{caption}
\usepackage[english]{babel}
\usepackage{blindtext}
\usepackage{tabularx}
\usepackage{threeparttable} 
\usepackage{subcaption}
\usepackage{comment}
\usepackage{makecell}
\usepackage{booktabs}
\usepackage{array}
\usepackage{multirow}

\usepackage{hyperref}
\usepackage{pifont}
\usepackage{amssymb}

\usepackage{stackengine,scalerel}

\usepackage{tikz}
\usepackage{xcolor}
\definecolor{semanticblue}{RGB}{0,122,204} 

\usepackage{diagbox} 
\usepackage{xcolor}
\definecolor{darkred}{RGB}{139,0,0}  
\definecolor{mypink}{RGB}{255,182,193} 
\definecolor{lightbrown}{RGB}{240,200,120} 
\definecolor{tealframe}{RGB}{0,166,162} 
\definecolor{customviolet}{RGB}{123,47,165} 

\newcommand{\halfcheck}{%
  \scalebox{0.9}{\ding{52}\hspace{-0.7em}\rotatebox[origin=c]{-9.2}{\ding{55}}}%
}

\newcommand{\revise}[1]{\textcolor{blue}{#1}}
\soulregister\cite7

\soulregister\eqref7

\begin{document}

\title{TripleMixer: A 3D Point Cloud Denoising Model for Adverse Weather}



\author{Xiongwei Zhao, \IEEEmembership{Graduate Student Member, IEEE}, Congcong Wen, \IEEEmembership{Member, IEEE}, Xu Zhu, \IEEEmembership{Senior Member, IEEE}, Yang Wang, Haojie Bai and Wenhao Dou

\thanks{This work was previously released as a preprint on arXiv.org (arXiv:2408.13802).}



\thanks{The datasets and code will be made publicly available at {\textcolor{red}{\textit{\url{https://github.com/Grandzxw/TripleMixer}}}}}

}


\maketitle

\begin{abstract}

Adverse weather conditions such as snow, fog, and rain pose significant challenges to LiDAR-based perception models by introducing noise and corrupting point cloud measurements. To address this issue, we propose TripleMixer, a robust and efficient point cloud denoising network that integrates spatial, frequency, and channel-wise processing through three specialized mixer modules. TripleMixer effectively suppresses high-frequency noise while preserving essential geometric structures and can be seamlessly deployed as a plug-and-play module within existing LiDAR perception pipelines. To support the development and evaluation of denoising methods, we construct two large-scale simulated datasets, Weather-KITTI and Weather-NuScenes, covering diverse weather scenarios with dense point-wise semantic and noise annotations. Based on these datasets, we establish four benchmarks: Denoising, Semantic Segmentation (SS), Place Recognition (PR), and Object Detection (OD). These benchmarks enable systematic evaluation of denoising generalization, transferability, and downstream impact under both simulated and real-world adverse weather conditions. Extensive experiments demonstrate that TripleMixer achieves state-of-the-art denoising performance and yields substantial improvements across all downstream tasks without requiring retraining. Our results highlight the potential of denoising as a task-agnostic preprocessing strategy to enhance LiDAR robustness in real-world autonomous driving applications.

\end{abstract}

\begin{IEEEkeywords}
Adverse Weather Dataset, Point Cloud Denosing, Robust LiDAR perception, Plug-and-Play Network
\end{IEEEkeywords}

\section{Introduction}

\begin{figure}[!t]
\centerline{\includegraphics[width=\columnwidth]{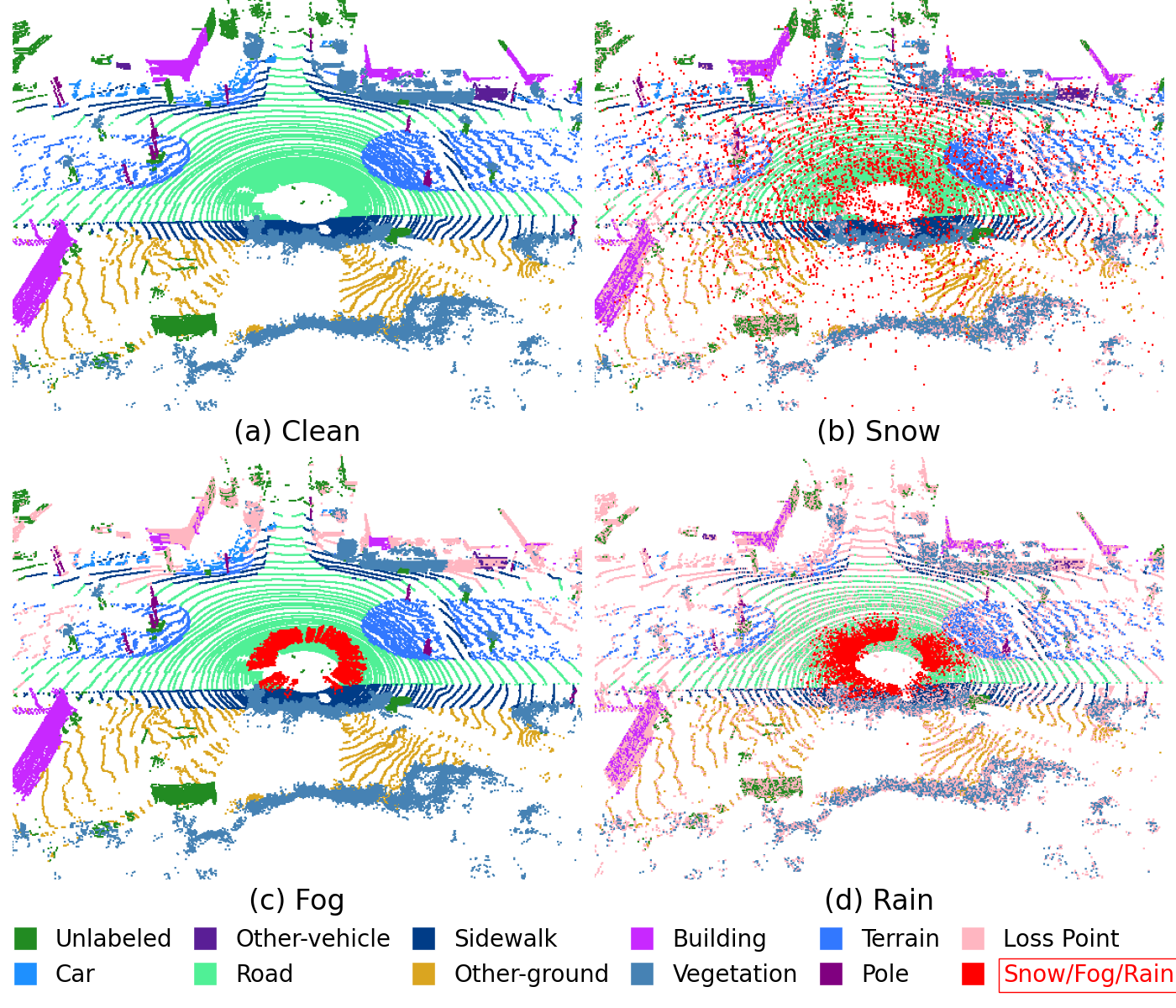}}
\caption{Visualization of semantic 3D points from the same LiDAR frame across four weather conditions in our proposed Weather-KITTI dataset. Weather noises are marked in \textcolor{red}{red} and loss points are marked in \textcolor{mypink}{pink}.} 
\vspace{-1.2em}
\label{distrubution}
\end{figure}

Robust LiDAR perception~\cite{kuang2025reslpr,zhao2024triplemixer} in adverse weather remains a critical challenge for autonomous driving systems. Weather-induced degradations such as snow, fog, and rain can cause LiDAR beams to scatter, reflect, or be absorbed abnormally, leading to corrupted point cloud measurements with substantial noise and missing points~\cite{safe}. These degradations compromise the reliability of 3D sensing and degrade the performance of downstream perception tasks~\cite{cai2021multi, wen2021airborne,wen2023pyramid,ma20233d,lin2025generalization}, including semantic segmentation, place recognition, and object detection.

To address this issue, recent research has explored two broad categories of solutions: domain transfer methods~\cite{li2023domain,li2024v2x,zhao2024unimix} and data preprocessing via denoising~\cite{jiang2024weather,liu2023dlc,3DOutDet}. Domain transfer methods, including domain adaptation and domain generalization techniques, aim to improve model robustness by adapting features across clean and adverse-weather domains. However, they typically require access to target domain data, involve task-specific retraining, and often lack generalizability across different perception tasks. In contrast, denoising serves as a task-agnostic preprocessing step that can be seamlessly integrated into existing LiDAR perception pipelines without the need for retraining.

Existing denoising approaches can be broadly categorized into statistical and learning-based methods~\cite{dreissig2023survey,surveyperception}. Statistical methods~\cite{sor,dsor,lior} rely on local geometric heuristics, but their iterative filtering procedures make them computationally inefficient and poorly scalable to large-scale point clouds. Learning-based methods typically follow one of two paradigms: projecting 3D point clouds onto 2D range images~\cite{wetahernet,4DenoiseNet} or directly processing raw point clouds~\cite{3DOutDet}. Most existing models exhibit limited ability to capture multi‑scale spatial structural information and lack frequency‑aware modeling to distinguish meaningful structures from high‑frequency noise, resulting in suboptimal denoising performance under adverse weather conditions.

To overcome these challenges, we propose \textit{TripleMixer}, a robust and efficient point cloud denoising architecture tailored for adverse weather conditions. TripleMixer consists of three complementary components: (1) a {Geometry Mixer (GMX)} that captures local geometric structures from raw 3D points while preserving fine-grained spatial details; (2) a {Frequency Mixer (FMX)} that performs multi-scale frequency decomposition across three orthogonal 2D projections to effectively separate structural information from high-frequency noise; and (3) a {Channel Mixer (CMX)} that facilitates inter-channel feature refinement to enhance contextual representation. This design enables TripleMixer to suppress noise while preserving critical geometric features, striking an effective balance between performance, efficiency, and interpretability.

To advance the research community’s capabilities in robust LiDAR perception under adverse weather, and to support the development and rigorous evaluation of TripleMixer, we construct two large-scale simulated datasets: \textit{Weather-KITTI} and \textit{Weather-NuScenes}. As shown in Table~\ref{overview}, our datasets contain over 200K LiDAR frames, substantially exceeding the scale of existing benchmarks, which typically include fewer than 50K frames. Moreover, our datasets preserve the characteristics of real LiDAR acquisitions while covering multiple adverse weather types (snow, fog, and rain). They also provide both point-wise semantic labels and weather annotations, which are lacking in most existing datasets. Built upon these datasets, we establish four benchmarks: (1) Denoising benchmark for evaluating denoising performance in isolation, and three downstream perception benchmarks under real-world adverse weather conditions: (2) Semantic Segmentation (SS), (3) Place Recognition (PR), and (4) Object Detection (OD). Our main contributions are summarized as follows:

\begin{itemize}

\item We propose TripleMixer, a plug-and-play point cloud denoising network that integrates spatial, frequency, and channel-wise processing through three specialized mixer layers. TripleMixer enables interpretable and robust denoising under adverse weather conditions, and can be seamlessly integrated into existing LiDAR perception pipelines to enhance their robustness.

\item  We introduce Weather KITTI and Weather NuScenes, two large-scale simulated LiDAR datasets encompassing diverse adverse weather conditions. Based on these datasets, we establish four benchmarks: Denoising, SS, PR, and OD. These benchmarks support comprehensive evaluation and provide a solid foundation for developing robust denoising and 3D perception models under both simulated and real-world adverse weather conditions.

\item Extensive experiments on both all benchmarks demonstrate that TripleMixer achieves state-of-the-art denoising performance. Furthermore, when used as a preprocessing module without retraining downstream models, TripleMixer significantly improves performance on SS, PR, and OD tasks under real-world adverse weather, achieving relative accuracy gains of up to 20\%, 56\%, and 16\%, respectively.

\end{itemize}

\section{Related Work}

\subsection{Robust LiDAR Perception}
Adverse weather conditions pose significant challenges to LiDAR-based perception by introducing severe noise and degradation into point cloud measurements~\cite{NDT}. Recent efforts to address this issue primarily fall into two categories: domain transfer and data preprocessing through denoising. 

Domain transfer methods include domain adaptation~\cite{li2023domain,xiao20233d,zhao2024unimix}, which leverages labeled or unlabeled target-domain (i.e., adverse weather) data to align feature distributions via retraining or fine-tuning, and domain generalization~\cite{fang2022weather,li2024v2x}, which enhances model robustness through data augmentation or domain-invariant representation learning. However, these approaches typically require access to target-domain data, involve task-specific retraining, and often struggle to generalize across diverse perception tasks.

In contrast, denoising offers a task-agnostic and modular preprocessing solution that can be seamlessly integrated into existing LiDAR perception pipelines. Existing denoising approaches can be broadly categorized into statistical and deep learning-based methods. Statistical methods, such as SOR and ROR~\cite{sor}, rely on local geometric heuristics, while dynamic variants like DROR~\cite{dror} and DSOR~\cite{dsor}, as well as intensity-aware filters such as LIOR~\cite{lior} and DDIOR~\cite{ddior}, offer improved resilience to weather-induced noise. However, their iterative nature makes them computationally expensive and limits scalability. Deep learning-based approaches learn complex noise patterns from data and are typically grouped by input modality: 2D range images or raw 3D point clouds. Range image-based methods, such as WeatherNet~\cite{wetahernet}, LiSnowNet~\cite{LiSnowNet}, and 4DenoiseNet~\cite{4DenoiseNet}, leverage CNNs or FFTs to process projected range representations, but inherently compromise the fidelity of 3D spatial structure. In contrast, point cloud-based models like 3D-OutDet~\cite{3DOutDet} directly operate on raw point clouds using neighborhood-based convolutions, preserving geometric detail but introducing significant runtime overhead due to $k$-NN preprocessing. Despite recent progress, existing methods remain limited in their capacity to capture multi-scale spatial structure and generally lack frequency-aware modeling, which is critical for distinguishing meaningful features from high-frequency noise.

\subsection{Adverse Weather Datasets and Benchmarks}

Existing adverse-weather datasets for LiDAR perception can be broadly categorized into two types: real-world datasets and simulated datasets. Real-world datasets are collected using actual sensors under varying environmental conditions, providing realistic examples of sensor behavior in adverse weather. For instance, the STF dataset~\cite{bijelic2020seeing} offers multimodal recordings with different levels of visibility, while the CADC dataset~\cite{CADC} captures winter driving scenes using a VLP-32 LiDAR and provides 3D bounding box annotations. The WADS dataset~\cite{dsor} focuses on extreme outdoor environments and includes point-wise annotations of weather-induced noise, making it suitable for fine-grained analysis.

Simulated datasets are typically created by applying weather simulators to existing clean-weather LiDAR datasets. These models simulate the physical interactions between LiDAR beams and environmental factors such as rain, fog, and snow. For example, a hybrid Monte Carlo model is employed in~\cite{lisa} to simulate raindrop interference, while fog effects are modeled using data-driven and physically based methods in~\cite{goodin2019predicting,fogsim}. Snow simulation is handled in~\cite{snowsim} by superimposing reflections from virtual particles. Building on these techniques, datasets such as SnowKITTI~\cite{4DenoiseNet} and Robo3D~\cite{robo3d} augment existing datasets like KITTI~\cite{kitti}, SemanticKITTI~\cite{Semantickitti}, and nuScenes~\cite{nuscenes} with weather effects. Some recent benchmarks~\cite{dong2023benchmarking,beemelmanns2024multicorrupt} also introduce synthetic perturbations across multiple sensor modalities to evaluate model robustness.

Overall, real-world datasets offer valuable realism by capturing authentic environmental variations and sensor responses. However, their scale is often constrained by the high cost of manual annotation, and they typically provide limited coverage across diverse weather conditions. In contrast, simulated datasets enable controllable generation of adverse weather effects and can effectively augment existing clean-weather datasets. Nonetheless, most existing simulated datasets are restricted in scale, lack fine-grained weather annotations, and have yet to be systematically validated for their effectiveness in enhancing downstream perception tasks under real‑world adverse weather conditions. To address these limitations, we introduce two large-scale simulated datasets and establish four benchmarks spanning both denoising and downstream LiDAR perception tasks under real-world adverse weather scenarios.

\begin{table*}[t]
\setlength{\tabcolsep}{4.0pt} 
\captionsetup{justification=centering, singlelinecheck=false}
\centering
\caption{Comparison of the Proposed Adverse Weather Dataset with Existing Public Datasets, including Year, LiDAR Parameters, Frame Count, Semantic Labels, Weather Types, Intensity Features, and Point-Wise Weather Noise Labels.}
\label{overview}
\begin{tabular}{lcccccccccc}
\toprule
\multirow{2}{*}{Dataset} & \multirow{2}{*}{Year} & \multirow{2}{*}{LiDAR} & \multirow{2}{*}{Size} & \multirow{2}{*}{Semantic Labels} & \multicolumn{3}{c}{Weather Types} & \multirow{2}{*}{Intensity} & \multirow{2}{*}{Weather Labels} \\  
 & & & &  &Snow  &Fog &Rain & \\
\midrule
WADS\cite{dsor} &2021  & 64 channel &1K frames  &point-wise &\checkmark &\ding{55}  &\ding{55}  &\checkmark & \checkmark \\  
CADC\cite{CADC}  &2021 & 32 channel &7K frames  &object-wise &\checkmark &\ding{55}  &\ding{55}  &\checkmark  &\ding{55}  \\ 
STF\cite{bijelic2020seeing}  &2020 & 32 \& 64 channel &13K frames &object-wise &\checkmark & \checkmark &\ding{55}   &\checkmark &\ding{55} \\ 
LIBRE\cite{carballo2020libre}  &2020 & 32 \& 64 \& 128 channel &6K frames &object-wise &\ding{55} &\checkmark &\checkmark   &\checkmark &\ding{55}\\ 
Boreas\cite{Boreas}  & 2023  & 128 channel  &7K frames  &object-wise &\checkmark  &\ding{55}  &\checkmark  &\checkmark &\ding{55} \\ 
SemanticSTF\cite{xiao20233d} &2023 & 64 channel &2K frames &point-wise &\revise{\checkmark} & \revise{\checkmark} &\revise{\checkmark}   &\revise{\checkmark} &\revise{\halfcheck}\\ 
\revise{Robo3d-KITTI\cite{robo3d}}  & \revise{2023}  & \revise{64 channel}  &\revise{8K frames} &\revise{object-wise} &\revise{\checkmark}  &\revise{\checkmark}  &\revise{\ding{55}} &\revise{\checkmark} &\revise{\ding{55}} \\ 
\revise{Robo3d-SemanticKITTI\cite{robo3d}}  & \revise{2023}  & \revise{64 channel}  &\revise{8K frames} &\revise{point-wise} &\revise{\checkmark}  &\revise{\checkmark}  & \revise{\ding{55}}  &\revise{\checkmark} &\revise{\checkmark} \\ 
\revise{Robo3d-nuScenes\cite{robo3d}} & \revise{2023} & \revise{32 channel} &\revise{12K frames} &\revise{point-wise}  &\revise{\checkmark}  &\revise{\checkmark}  &\revise{\ding{55}}   &\revise{\checkmark} &\revise{\ding{55}}  \\ 
\revise{Robo3d-WOD\cite{robo3d}} & \revise{2023} & \revise{64 channel} &\revise{-} &\revise{-}  &\revise{\checkmark}  &\revise{\checkmark}  &\revise{\ding{55}}   &\revise{\checkmark} &\revise{-}  \\ 
SnowKITTI\cite{4DenoiseNet}  & 2023  & 64 channel  &43K frames  &\ding{55} &\checkmark   &\ding{55}  &\ding{55}  &\ding{55} &\checkmark\\ 
\hline
\textbf{Weather-KITTI (Ours)} &  2024 & 64 channel &130K frames  &point-wise  &\checkmark & \checkmark & \checkmark & \checkmark &\checkmark \\
\textbf{Weather-NuScenes (Ours)} & 2024 & 32 channel &84K frames  & point-wise  &\checkmark & \checkmark & \checkmark & \checkmark &\checkmark \\  \hline
\end{tabular}

\begin{tablenotes}
\small  
\item \halfcheck: SemanticSTF provides weather‑related labels for indiscernible points under adverse weather but lacks separate noise annotations, \par 

\checkmark: included,\quad \ding{55}: not included,\quad –: unknown, 

\end{tablenotes}
\vspace{-1.0em}
\end{table*}


\section{Weather‑KITTI and Weather‑NuScenes Datasets}
As introduced earlier, we propose our simulated adverse weather datasets, named \textbf{Weather-KITTI} and \textbf{Weather-NuScenes}, which are based on the SemanticKITTI\cite{Semantickitti} and nuScenes-lidarseg\cite{nuscenes} datasets, respectively. To simulate three adverse weather conditions: snow, fog, and rain, we employ the LiDAR Snowfall Simulation (LSS) \cite{snowsim} for snow, the LiDAR Fog Simulation (LFS) \cite{fogsim} for fog, and the LiDAR Light Scattering Augmentation (LISA) \cite{lisa} for rainy conditions. These simulation methods produce realistic weather effects, and they have demonstrated that perception models trained on their simulated data exhibit improved robustness under adverse weather conditions. Table \ref{overview} provides a detailed comparison between the proposed dataset and existing publicly available adverse‑weather datasets. Compared to these datasets, our dataset offers a larger scale, covers a broader range of weather scenarios, preserves complete point cloud information, and provides dense point‑wise semantic and weather annotations.

\begin{figure*}[t]
	\centering
	\captionsetup[subfigure]{margin=0.4pt} 
	\subfloat[\footnotesize{Weather-KITTI}]{\includegraphics[width = 0.5\textwidth]{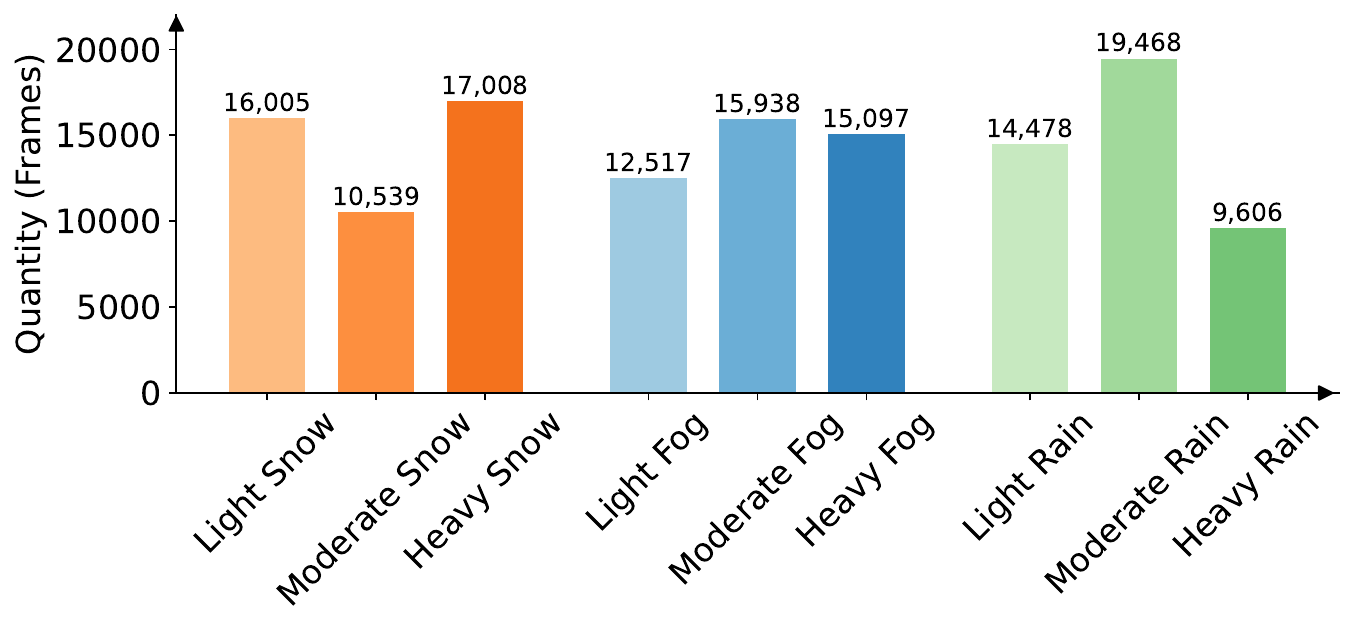}}
	\hfill
	\subfloat[\footnotesize{Weather-NuScenes}]{\includegraphics[width = 0.5\textwidth]{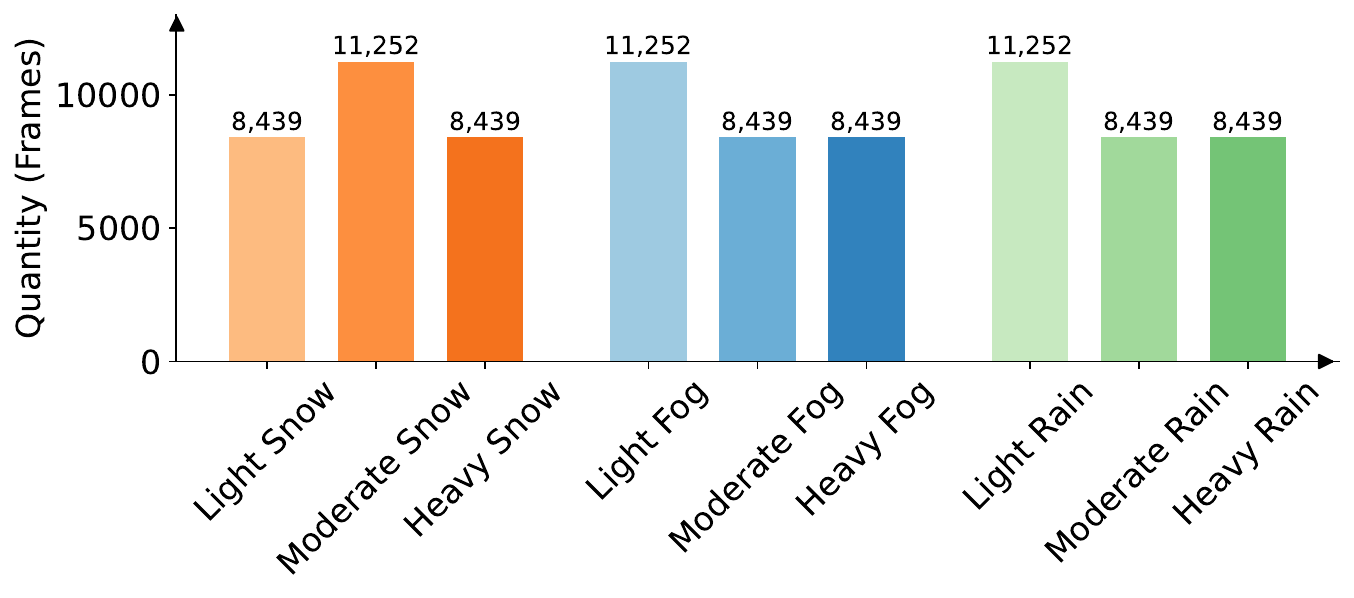}} 
 
	\caption{{LiDAR data frames for different levels of weather severity in our proposed Weather-KITTI and Weather-NuScenes datasets. LiDAR frames corresponding to different weather severities are randomly selected.} }
	\label{Weather}
 \vspace{-1.2em}
\end{figure*}

\begin{figure*}[t]
    \centering
    \captionsetup[subfigure]{margin=0.4pt}
    
    \subfloat[\footnotesize{Semantic Distribution - Weather-KITTI}]
    {\includegraphics[width = 0.5\textwidth]{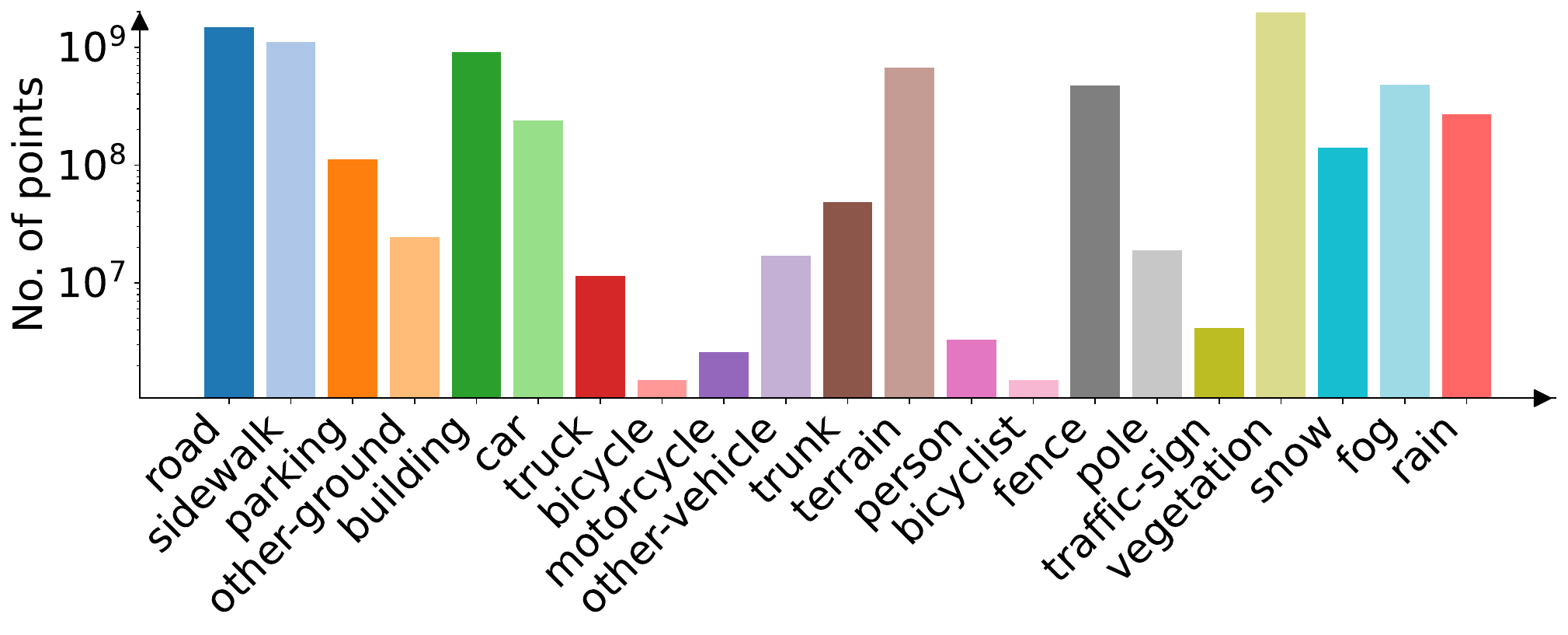}}
    \hfill
    \subfloat[\footnotesize{Semantic Distribution - Weather-NuScenes}]
    {\includegraphics[width = 0.5\textwidth]{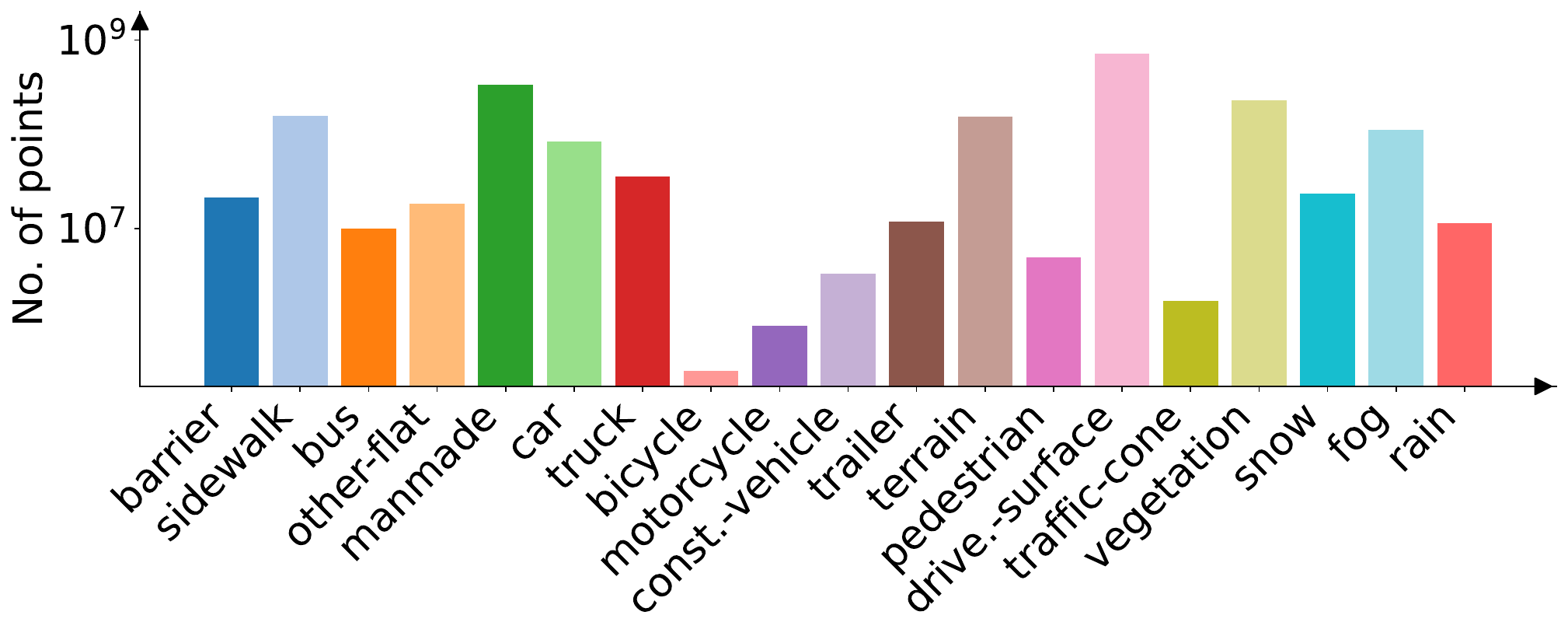}}
    
    \vspace{1ex}
    
    \subfloat[\footnotesize{Weather Impact per Semantic Class - Weather-KITTI}]
    {\includegraphics[width = 0.5\textwidth]{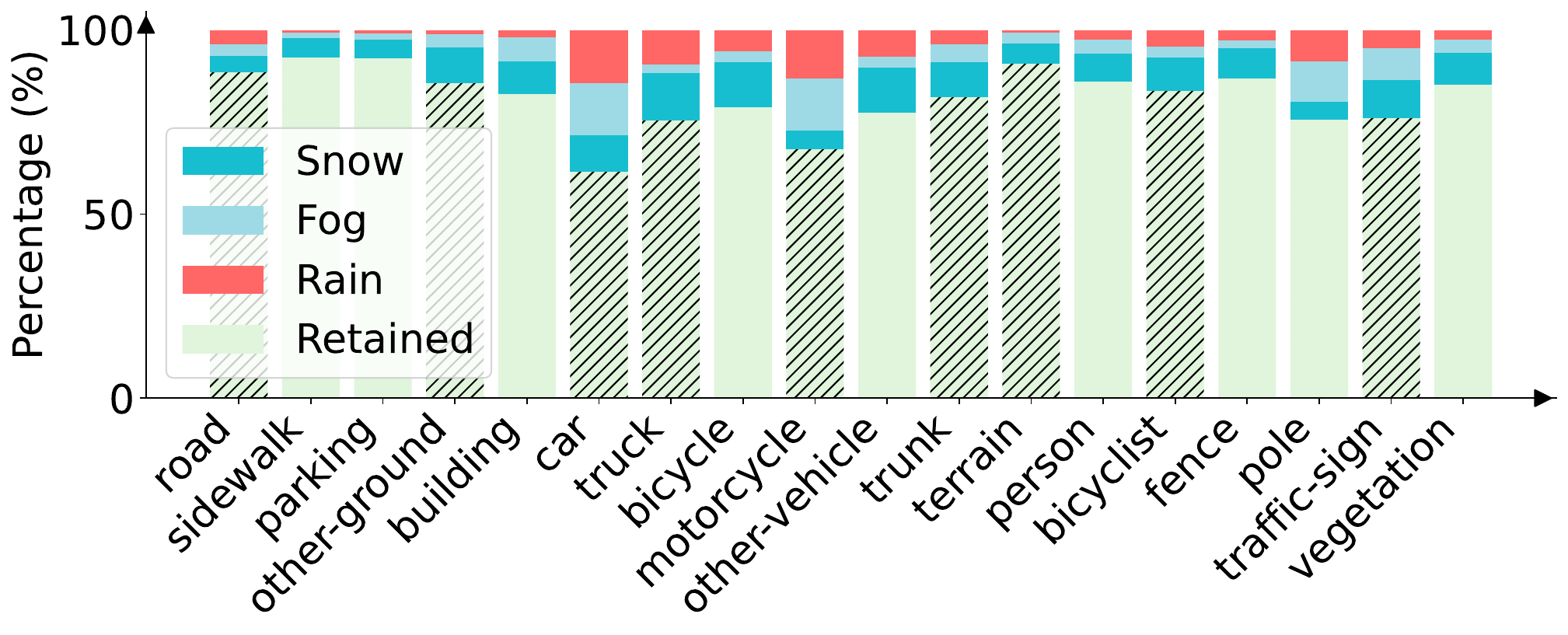}}  
    \hfill
    \subfloat[\footnotesize{Weather Impact per Semantic Class - Weather-NuScenes}]
    {\includegraphics[width = 0.5\textwidth]{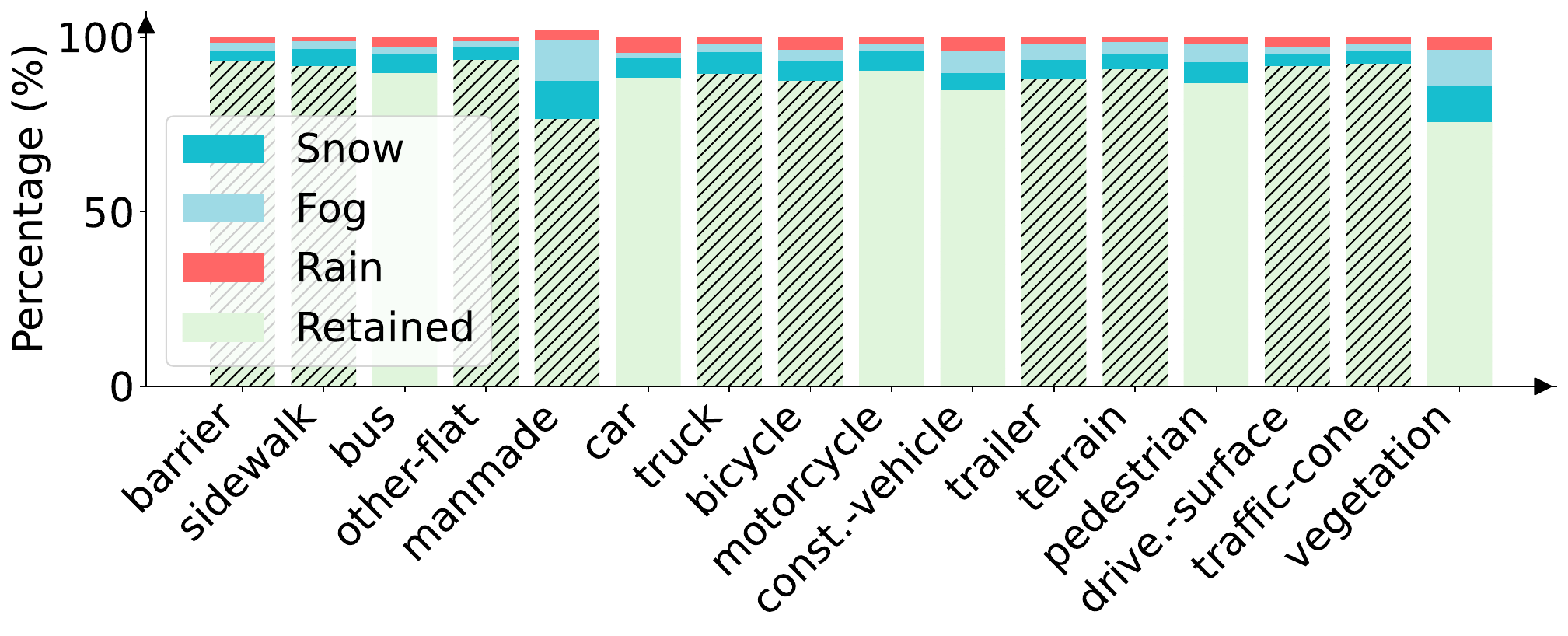}}  

    \caption{Top: Semantic distribution of 3D points for all sequences in Weather-KITTI and Weather-NuScenes. 
    Bottom: Distribution of weather-affected points across semantic classes under snow, fog, and rain. “Retained” denotes the percentage of points in that class unaffected by weather artifacts.}
    \label{weather_class_distribution}
    \vspace{-1.2em}
\end{figure*}

\begin{table}[!t]
\captionsetup{justification=centering, singlelinecheck=false}
\centering
\caption{{Details of Weather Simulations.}}
\label{simulations_config}
\begin{tabular}{ccc}
\toprule
\textbf{Weather} & \textbf{Severity} & \textbf{Description} \\
\midrule
\multirow{3}{*}{Snowfall Simulation} & Light & snowfall rate \(r_{s} = [0.5, 1.0]\) mm/h \\
 & Moderate & snowfall rate \(r_{s} = [1.5, 2.0]\) mm/h \\
 & Heavy & snowfall rate \(r_{s} = [2.5, 3.0]\) mm/h \\
\midrule
\multirow{3}{*}{Fog Simulation} & Light & fog simulation with \(\beta = [0.01,0.05]\) \\
 & Moderate & fog simulation with \(\beta = [0.08, 0.14]\) \\
 & Heavy & fog simulation with \(\beta = [0.18,0.25]\) \\
\midrule
\multirow{3}{*}{Rain Simulation} & Light & rain rate \(r_{r} = [1.0, 1.5]\) mm/h \\
 & Moderate & rain rate \(r_{r} = [1.8, 2.4]\) mm/h \\
 & Heavy & rain rate \(r_{r} = [2.6, 3.0]\) mm/h \\
\bottomrule
\end{tabular}
\vspace{-1.2em}
\end{table}

\subsection{Dataset Statistics}
Figure~\ref{distrubution} shows the semantic distributions of a single LiDAR frame captured under three different weather conditions from our Weather‑KITTI dataset, along with the corresponding clean reference frame. Weather‑induced noise points are predominantly concentrated near the LiDAR sensor, with snow producing more dispersed and irregular noise patterns than rain and fog. Table \ref{simulations_config} presents the severity levels of various weather-induced corruptions configured in this study. Based on Table \ref{simulations_config}, we simulated different severity levels (light, moderate, and heavy) of snow, fog, and rain.
As weather severity increases, the degree of LiDAR point cloud corruption escalates accordingly. All LiDAR frames in the Weather‑KITTI and Weather‑NuScenes datasets were randomly synthesized across the three severity levels, with a comparable number of frames maintained for each level to ensure balanced data distribution. As illustrated in Figure~\ref{Weather}(a), Weather-KITTI contains approximately 14K frames per weather type at each severity level, while Weather-NuScenes includes around 9K frames per category and severity level, as shown in Figure~\ref{Weather}(b). In Weather‑KITTI, the proportions of degraded points at the light, moderate, and heavy severity levels across all weather classes (snow, fog, rain) are 17.55\%, 30.85\%, and 51.60\%, respectively; In Weather‑NuScenes, the corresponding values are 14.51\%, 37.21\%, and 48.28\%.

Weather‑KITTI, based on SemanticKITTI~\cite{Semantickitti}, comprises 22 sequences with 130,656 LiDAR scans. Each point cloud is segmented into 21 semantic classes, with the weather-related classes for snow, fog, and rain mapped to labels 110, 111, and 112, respectively. As SemanticKITTI provides complete semantic labels only for the first ten sequences, we restrict the use of standard semantic labels accordingly. However, our dataset provides weather class annotations across all 22 sequences. Weather-NuScenes is derived from nuScenes-lidarseg~\cite{nuscenes} and includes 84,390 LiDAR scans evenly distributed across ten sequences. It covers 19 semantic classes, along with added labels for snow, fog, and rain using the same indices as in Weather-KITTI. The top row of Figure~\ref{weather_class_distribution} shows the total number of LiDAR points per semantic class for all sequences in our datasets: Weather-KITTI contains 139 million, 383 million, and 89 million points corrupted by snow, fog, and rain noise, respectively. Similarly, Weather-NuScenes includes 23 million, 112 million, and 6 million corrupted points under the same conditions. The bottom row of Figure~\ref{weather_class_distribution}  presents the proportion of points affected by each weather artifacts per semantic class. Weather‑KITTI exhibits more pronounced weather effects on traffic‑related categories such as cars, trucks, and motorcycles, which are typically located near the LiDAR sensor and in central road regions where particles cause stronger occlusion, scattering, and reflection. In contrast, Weather‑NuScenes shows a more balanced distribution of weather corruption across classes, reflecting its diverse urban scenes and multiple LiDAR viewpoints.

\begin{figure}[t]
    \centering
    \captionsetup[subfigure]{margin=0.4pt}
    
    \subfloat[\footnotesize{WADS}]
    {\includegraphics[width = 0.5\columnwidth]{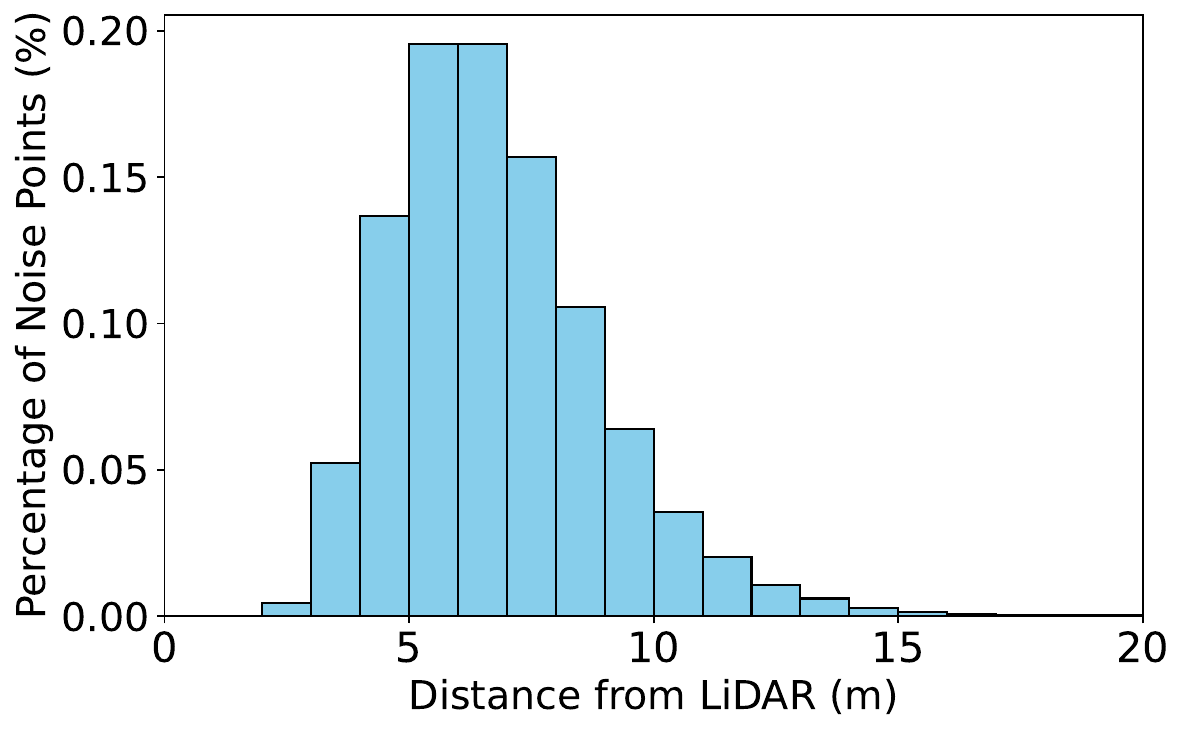}}
    \hfill
    \subfloat[\footnotesize{Ours}]
    {\includegraphics[width = 0.5\columnwidth]{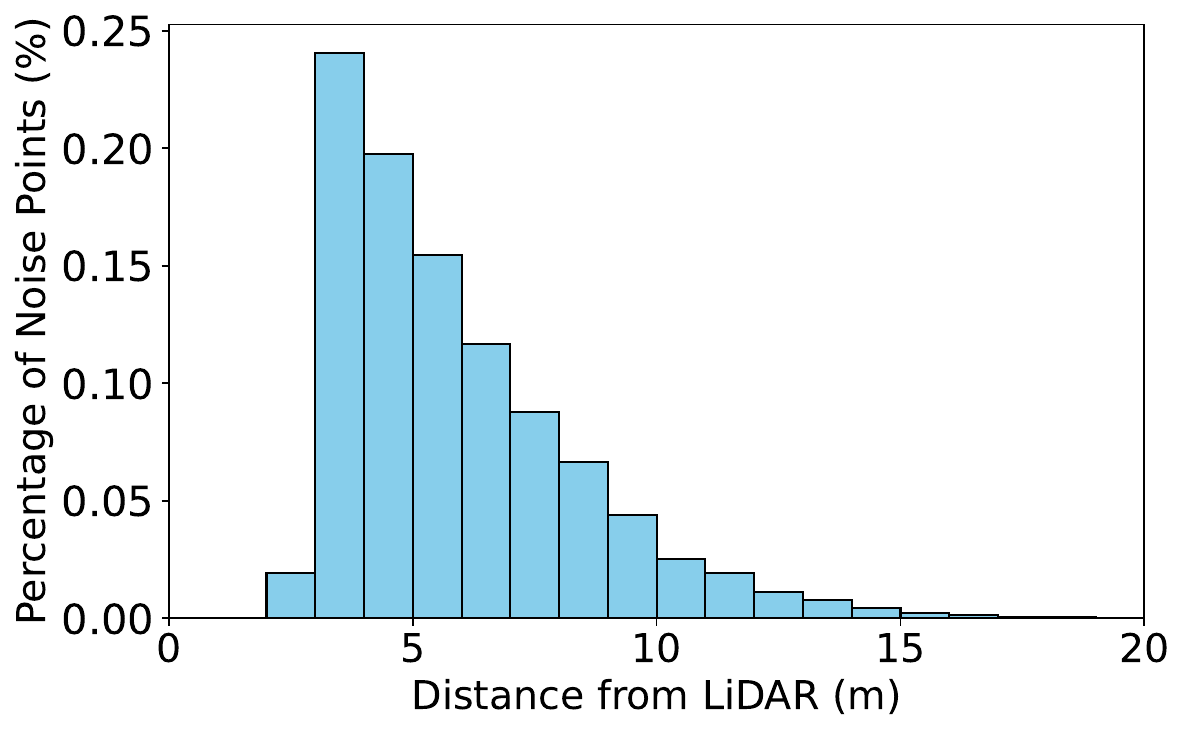}}
    \caption{Spatial Distribution Characteristics of Noise Points in WADS and Our Dataset.}
    \label{spatial_distribution}

\end{figure}

\begin{table}[t]
\centering
\caption{Intensity Distribution Characteristics of Noise Points in WADS and Our Dataset.}
\label{intensity_dist}
\begin{tabular}{ccc}
\toprule
\textbf{Intensity Range} & \textbf{WADS (\%)} & \textbf{Ours (\%)} \\
\midrule
$\left[0.0,\ 0.3\right]$ & 96.63 & 91.92 \\
$\left[0.3,\ 0.6\right]$ & 3.01 & 7.54 \\
$\left[0.6,\ 1.0\right]$ & 0.36 & 0.54 \\
\bottomrule
\end{tabular}
\vspace{-1.0em}
\end{table}

\subsection{Dataset Comparison}
To assess the practical reliability of our simulated datasets, we compare them with WADS~\cite{dsor}, which provides real‑world LiDAR sequences collected in snowy conditions with annotated noise labels. We randomly sample 1,000 scans from each dataset and analyze two key noise properties: spatial and intensity distributions. As illustrated in Figure~\ref{spatial_distribution}, both datasets exhibit similar spatial characteristics. Approximately 90\% of noise points are concentrated within a range of 3 to 10 meters from the LiDAR sensor, with a gradual decline beyond 10 meters and negligible presence past 20 meters. This consistency indicates that our simulations effectively reproduce real‑world occlusion and backscatter effects. Moreover, the intensity distribution of noise points, summarized in Table~\ref{intensity_dist}, further demonstrates the similarity. Over 90\% of noise points in both datasets fall within the low-intensity range [0.0, 0.3], indicating consistent degradation characteristics. Minor differences in the mid-range interval [0.3, 0.6] may be attributed to variations in LiDAR hardware or scene-specific factors, yet the overall distribution trend remains well aligned. These findings further validate that our synthetic weather corruption effectively replicates real-world noise patterns in both spatial and intesity properties.

\begin{figure*}[t]
        \centerline{\includegraphics[width=\linewidth]{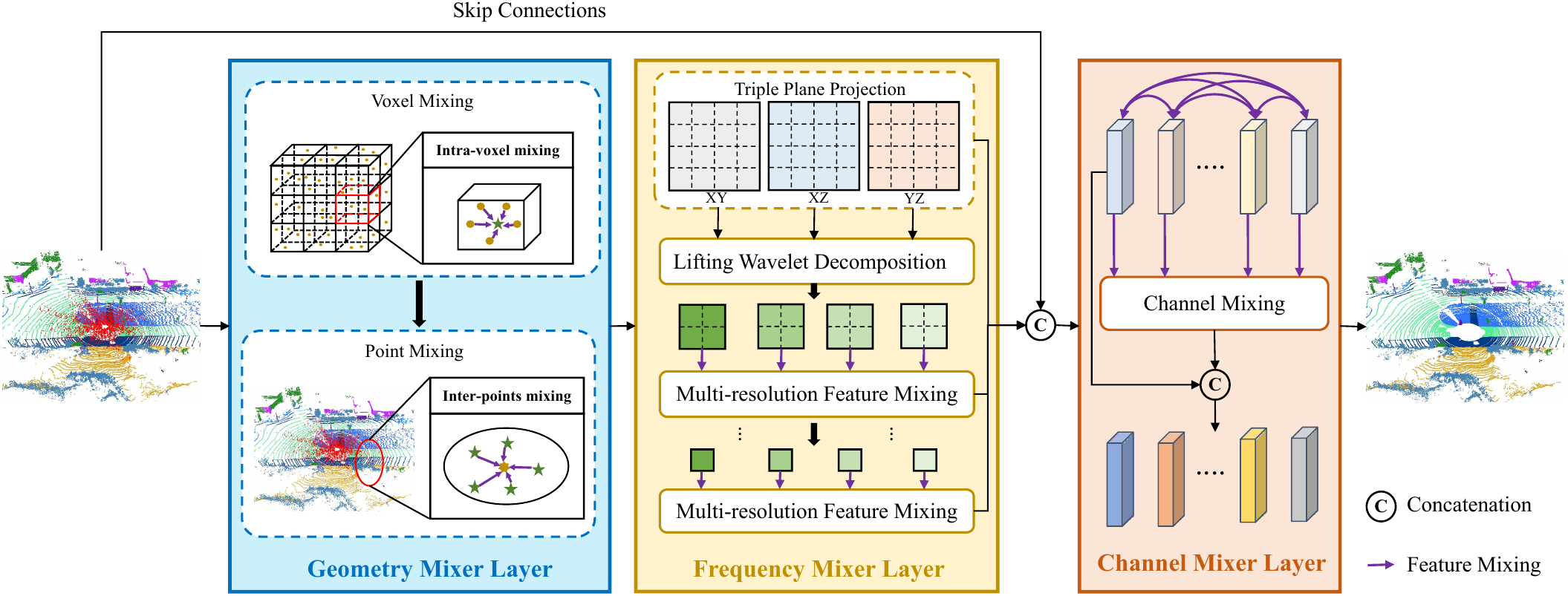}}
	\caption{Overview of the proposed TripleMixer denoising network, which consists of three key layers: the Geometry Mixer (GMX), Frequency Mixer (FMX), and Channel Mixer (CMX). \revise{TripleMixer is trained solely  using point-wise weather labels.}}
	\label{ov}
    \vspace{-1.0em}
\end{figure*}

\vspace{1.2em}

\section{TripleMixer Model}
\subsection{Problem Statement}
In this study, we aim to perform adverse weather denoising from LiDAR point clouds to restore clean point clouds. Given an adverse weather dataset of $N$ LiDAR scans $\left\{\left(P_{i}, L_{i}\right) \mid i=1, \ldots, N\right\}$, where $P_{i}\in\mathbb{R}^{n_{i}\times5}$ is the $i$th point set containing $n_i$ LiDAR points. Each row of $P_{i}$ consists of five features representing one LiDAR point $p$, namely ($x$, $y$, $z$, $i$, $r$). ($x$, $y$, $z$) denote the Cartesian coordinates of the point relative to the LiDAR, $i$ indicates the intensity of the returning laser beam and $r$ represents the 3D spatial distance from the point to the LiDAR center. $L_{i}\in\mathbb{Z}^{n_{i}}$ contains the ground truth labels for each point $p$ in $P_{i}$. Our objective is to learn a denoising model $\Phi$ that outputs a label set $\hat{L_{i}}\in\mathbb{Z}^{n_{i}}$, designed to remove outliers from the point cloud data as follows:
\begin{equation}
\hat{L_{i}}=\Phi({P_{i}} ; {\theta}) ,
\end{equation}
where $\theta$ denotes the model parameters to be optimized, aiming to minimize the difference between the prediction $\Phi({P_{i}} ; {\theta})$ and the ground truth labels $L_{i}$.

\subsection{Overall Framework} 
In this section, we propose the TripleMixer, a robust denoising model designed for removing 3D point cloud noises under adverse weather conditions. Its architecture is illustrated in Figure~\ref{ov} and comprises three primary layers: 1) the Geometry Mixer (GMX) Layer, which mixes the spatial geometric relationships of neighboring points, thereby enabling the network to discern crucial local structural features and maintain important geometric details; 2) the Frequency Mixer (FMX) Layer, employing multi-scale wavelet analysis to achieve a more comprehensive representation of both noise and normal points in the frequency domain; and 3) the Channel Mixer (CMX) Layer, which enhances the overall feature representation by mixing contextual information across channels, improving the model's accuracy and adaptability in complex environments. Specifically, in the GMX layer, we first voxelize the LiDAR point cloud in 3D space to facilitate downsampling, then select K neighboring points and employ the attentive pooling operation to effectively capture local geometric patterns. For the FMX layer, we initially quantize the 3D features derived from the GMX layer along the X, Y, and Z axes into 2D grids. These grids are projected onto the YZ, XZ, and XY planes, creating the corresponding 2D grid images for triple plane projection. Subsequently, these images are processed using an adaptive wavelet module that provides a multi-resolution decomposition, efficiently enhancing the learning of multi-scale features. In the CMX layer, we mix feature information across different channels, synthesizing information across various directions and scales, thereby enriching the network with superior contextual information. Detailed discussions of these components will follow in subsequent sections.

\begin{figure}[!t]
	\centerline{\includegraphics[width=\columnwidth]{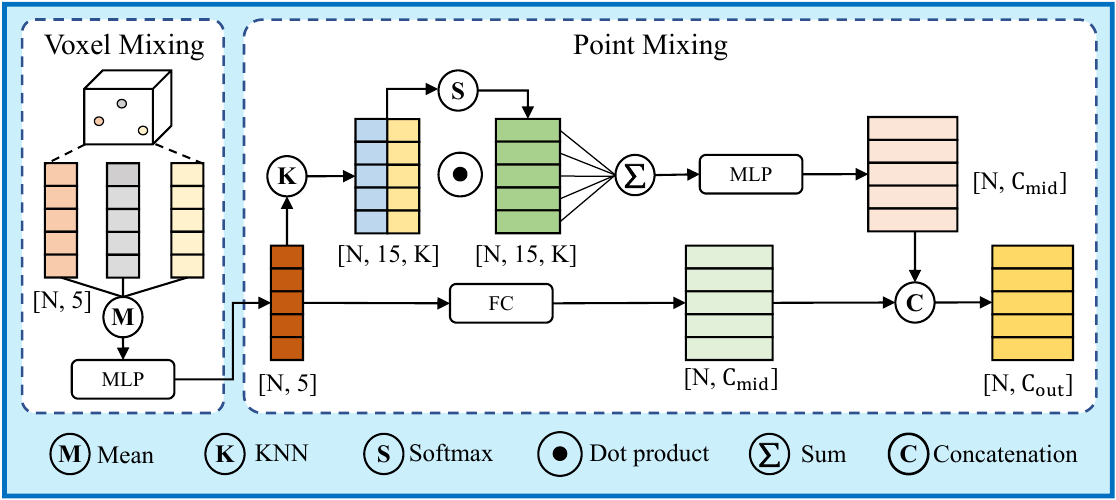}}
	\caption{Details of the Geometry Mixer Layer. The input features within the same voxel are meaned (downsampled), followed by the KNN neighborhood feature encoding, an attentive pooling operation, and a residual connection to obtain the output features.} 
	\vspace{-1.0em}
	\label{gmx}
\end{figure}

\begin{figure*}[ht]
	\centerline{\includegraphics[width=0.95\linewidth,height = 0.26\textheight]{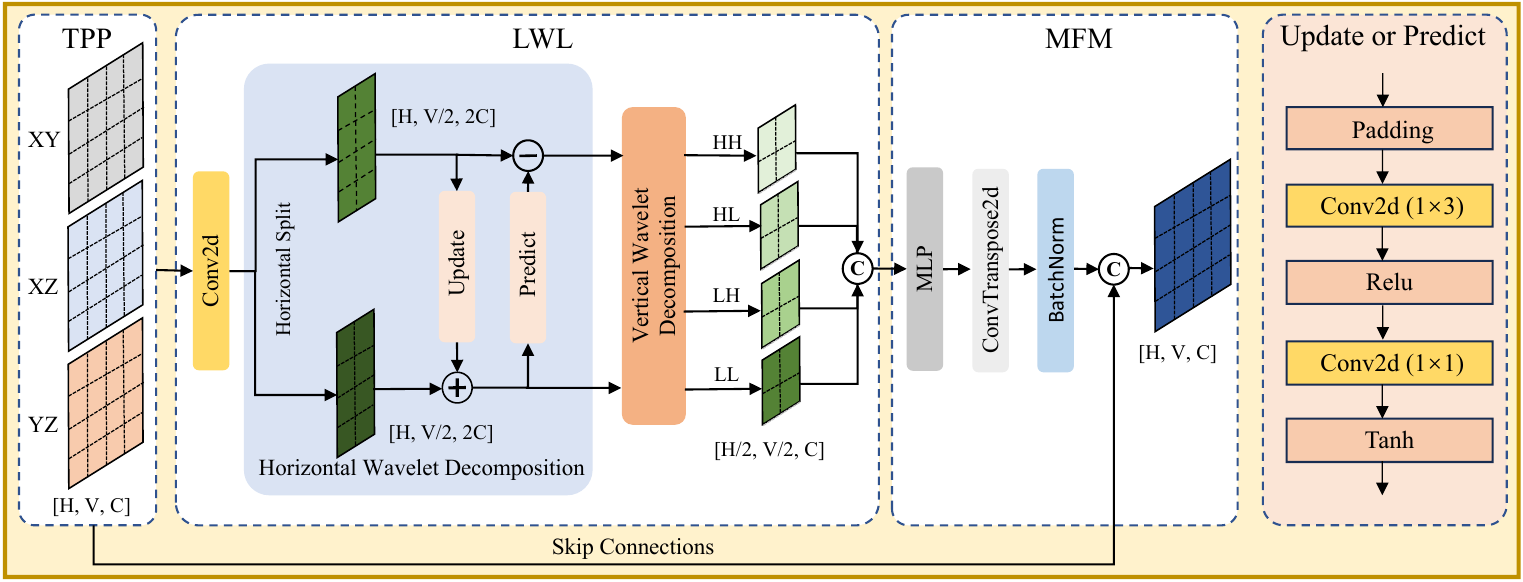}}
	\caption{Architecture of the Frequency Mixer Layer. The Frequency Mixer Layer consists of three modules: the Triple Plane Projection (TPP) Block, the Lifting Wavelet (LWL) Block, and the Multi-resolution Feature Mixing (MFM) Block. It accepts input features with dimensions [H, V, C], where H and V represent height and width, and C represents the number of channels.}
	\label{wave}
 \vspace{-1.2em}
\end{figure*}

\subsection{Geometry Mixer (GMX) Layer}
The GMX layer is a pivotal component of our TripleMixer, which efficiently encodes and aggregates local geometric features, enhancing the model's ability to analyze 3D point cloud. By mixing spatial information at the intra-voxel level and geometric information at the inter-point level, the GMX layer captures the spatial structures and relationships around each point, thereby forming a richer and more discriminative feature representation. The structure of the GMX layer is shown in Figure \ref{gmx}. The GMX layer includes Voxel Mixing and Point Mixing. Initially, Voxel Mixing is performed by averaging (downsampling) all point features within the same voxel, followed by an Multi-Layer Perceptron (MLP) to introduce non-linearity, resulting in a unique center point $p_{i}\in\mathbb{R}^{5}$ that best represents the voxel. Then, for each of the nearest K points $\left\{p_{i}^{1}, \cdots p_{i}^{k}, \cdots p_{i}^{K}\right\}\in\mathbb{R}^{5}$ of a center point $p_{i}$, we explicitly perform local point geometric feature mixing $l_{i}^{k}$ as follows:
\begin{equation}
l_{i}^{k} = MLP\left(p_{i} \oplus p_{i}^{k} \oplus\left(p_{i}-p_{i}^{k}\right)\right),
\end{equation}
where $\oplus$ is the concatenation operation, and $p_{i}-p_{i}^{k}$ represents the geometric feature similarity between the center point $p_{i}$ and the k-th neighboring point $p_{i}^{k}$, the MLP effectively models the spatial relationships between these points by learning from the concatenated features and their geometric similarities. Subsequently, we apply attentive pooling operation to automatically learn and emphasize important local features:
\begin{equation}
c_{i}^{k} = Softmax\left(\left(FC(l_{i}^{k})\right)\right) ,
\end{equation}
\begin{equation}
f_{i} = \sum_{i=1}^{K} c_{i}^{k} \odot l_{i}^{k} ,
\end{equation}
where $c_{i}^{k}$ is the unique attention score for the feature $l_{i}^{k}$, and $\odot$ denotes the dot product. In Eq. (13) and Eq. (14), a fully connected layer (FC) is used for point feature encoding, and the Softmax function is employed to learn the score $c_{i}^{k}$. These scores are then used to weight the features, which are subsequently summed to generate a new informative feature $f_{i}$. Finally, to prevent overfitting and preserve foundational geometric features, a residual connection is utilized to integrate the original point geometric features $p_{i}$ with the important local features $f_{i}$ obtained from the attentive pooling. The final output features of the GMX layer is computed as: 
\begin{equation}
F_{i} = Max\left(\left(MLP(f_{i})\right)\right) \oplus p_{i} .
\end{equation}

\vspace{-1.2em}
\subsection{Frequency Mixer (FMX) Layer}
In this section, we introduce the Frequency Mixer (FMX) Layer that designs a denoising lifting wavelet module to capture multi-scale geometric representation in the frequency domain, with its architecture is illustrated in Figure \ref{wave}. Noise in point clouds is usually randomly distributed in the spatial domain, manifesting as high-frequency components in the frequency domain, while the main structure of the original point cloud is represented by low-frequency components in the frequency domain. Our proposed denoising lifting wavelet module can decompose the features into different scale sub-bands, each containing various frequency information. By integrating these sub-bands, our model can effectively utilize both high-frequency and low-frequency components, resulting in a comprehensive and robust feature representation. 

The FMX layer consists of three blocks: \textbf{1) the triple plane projection block}, which projects the 3D point cloud onto 2D images. This block simplifies the inherent complexity of processing 3D point clouds while ensuring that sufficient spatial information is retained. \textbf{2) the lifting wavelet block}, which performs a hierarchical decomposition of each 2D images into four sub-band components. This block efficiently generates both approximation and detail coefficients, capturing both low-frequency and high-frequency information at each scale. \textbf{3) the multi-resolution feature mixing block}, which mixes decomposed components from multiple scales and combine this information across multiple resolutions to significantly enhance feature representation and analysis. 

\textbf{Triple Plane Projection (TPP) Block.} It aims to obtain the 2D features $x_{i}$ of the triple projection planes from the 3D features $F_{i}$ outputted by the GMX layer. To address reduce the computational complexity of 3D convolution operations and the loss of height-related details in BEV projection, we introduce a novel approach by projecting the 3D point cloud along the X, Y, and Z axes onto three distinct 2D planes at varying resolutions, as illustrated in Figure \ref{wave}. The TPP block utilizes sparse tensor operations to efficiently handle data. It starts by identifying non-zero indices in the point cloud to create sparse projection matrices for quantizing 3D points into 2D planes. The value of each pixel in the 2D plane represents the average of the 3D features falling within the same pixel. This triple plane projection method reduces redundancy and computational load while capturing detailed object features and enhancing spatial information, leading to more precise and accurate object characteristics.

\textbf{Lifting Wavelet (LWL) Block.} It consists of a horizontal wavelet decomposition (HWD) followed by two independent vertical wavelet decomposition (VWD) that generate the four sub-bands of the input 2D feature from the TPP block at each scale, as illustrated in Figure \ref{wave}. Suppose that the input 2D feature at decomposition layer $t$ is $x^{t}_{i}\in\mathbb{R}^{H \times V \times C}$, the LWL block first splits the input feature $x^{t}_{i}$ into two non-overlapping partitions along the horizontal direction, divided into even and odd components denoted as $x^{t}_{e}\in\mathbb{R}^{H \times V/2 \times C}$ and $x^{t}_{o}\in\mathbb{R}^{H \times V/2 \times C}$, respectively. The horizontal wavelet decomposition yields the detail components $d^{t}$ and the approximation components $c^{t}$, the details components $d^{t}$ represent the difference between $x^{t}_{o}$ and a predictor operator $P\left(\cdot\right)$ on $x^{t}_{e}$. Conversely, the approximation components $c^{t}$ are computed as $x^{t}_{e}$ plus an update operator $U\left(\cdot\right)$ on $d^{t}$. The process of the horizontal wavelet decomposition is described as follows:
\begin{equation}
d^{t}=x^{t}_{o}-P\left(x^{t}_{e}\right) ,
\end{equation}
\begin{equation}
c^{t}=x^{t}_{e}+U\left(d^{t}\right) ,
\end{equation}
where the structure of both the $P(\cdot)$ and $U (\cdot)$ operators is shown in Figure \ref{wave}. The operation for each begins with reflection padding to keep the original sequence length. This is followed by a $1\times3$ convolutional layer to process the features, activated by ReLU for non-linearity. Subsequently, a $1\times1$ convolution then adjusts the channels, and a Tanh activation function retain negative values in the output. 

Next, two independent VWD module, which have a structure similar to the HWD module, are respectively applied to the approximation components $c^{t}$ and the details components $d^{t}$ obtained from the HWD. Consequently, the four sub-band components $\left[{x^{t}_{HL},x^{t}_{HH},x^{t}_{LL},x^{t}_{LH}}\right] \in\mathbb{R}^{H/2 \times V/2 \times C}$ at layer $t$ can be obtained as follows:
\begin{equation}
[x^{t}_{HL},x^{t}_{HH}]= \mathit{VWD}\left(d^{t}\right) ,
\end{equation}
\begin{equation}
[x^{t}_{LL},x^{t}_{LH}]= \mathit{VWD}\left(c^{t}\right) ,
\end{equation}
where $L$ and $H$ denote low and high frequency information, respectively, while the first and second positions in each sub-band component' indices correspond to the horizontal and vertical directions, respectively. Finally, we concatenate the four sub-bands along the channel dimension, obtaining the output feature $x^{t}_{w}$ of the LWL block:
\begin{equation}
x^{t}_{w} = x^{t}_{LL} \oplus x^{t}_{LH}  \oplus x^{t}_{HL} \oplus x^{t}_{HH} ,
\end{equation}
where $x^{t}_{w}\in\mathbb{R}^{H/2 \times V/2 \times 4C}$. Unlike pooling or strided convolutions, the LWL block is information lossless, as it expands the number of channels by the same factor by which it reduces spatial resolution. The low frequency sub-band $x^{t}_{LL}$ serves as the input feature for the lifting wavelet decomposition in the next layer $t+1$. This recursive process is repeated at each scale, capturing multi-scales geometric information effectively. The maximum number of scales can be determined analytically based on the initial resolution of the triple projection planes, as detailed in Section\uppercase\expandafter{\romannumeral6-A}.

\textbf{Multi-resolution Feature Mixing (MFM) Block.} It comprises an MLP layer, a transposed convolution (TConv), a batch normalization (BN) layer, and residual connections. The concatenated feature $x^{t}_{w}$ from the LWL block is sequentially processed through these modules, as detailed below:
\begin{equation}
W^{t}_{i} = BN\left(TConv\left(MLP(x^{t}_{w})\right)\right) \oplus x^{t}_{i} ,
\end{equation}
where $x^{t}_{w}$ is initially processed through an MLP to restore the original channel dimension and mix multi-resolution features in the frequency domain. Subsequently, transposed convolutions are employed to upsample the feature maps, restoring the spatial resolution of the sub-band feature maps and providing access to both local and global information. The batch normalization (BN) layer normalizes the feature maps, stabilizing the training process and enhancing training stability and generalization. Finally, a residual connection reintroduces the original input feature $x^{t}$ at layer $t$, ensuring that while complex features are extracted, essential information from the initial input is retained. This process is repeated at multiple scales within the FMX layer, recursively mixing and enhancing multi-scale geometric representations.

\subsection{Channel Mixer (CMX) Layer}
The CMX layer plays a crucial role in reprojecting the 2D features processed by the FMX layer back into the 3D space, mixing contextual information across channels, and enhancing the overall feature representation. The structure of the CMX layer is shown in Figure \ref{ov}. This layer is essential for integrating multi-scale geometric features $W_{i}$ from the FMX layer are effectively combined and utilized for subsequent point denoising tasks. The processing flow of the CMX layer is as follows:
\begin{equation}
C_{i} = Drop\left(GConv\left(MLP\left(BN(W_{i})\right)\right)\right) \oplus x_{i} ,
\end{equation}
where $Drop$ represents the dropout operation to prevent overfitting. $GConv$ represents the group convolution used to mix channel feature information. Initially, the input feature $W^{t}$ undergoes the BN operation. Then, an MLP layer enhances dimensionality, followed by a group convolutional layer for mixing contextual information. A dropout operation prevents overfitting. Finally, a residual connection reintroduces the original input feature $x_{i}$ across the channels, capturing and fusing local-to-global context.

\subsection{Loss Function}
To train our proposed model and ensure that the captured information is relevant to the point cloud denoising task, the overall loss function consists of three components: the cross-entropy loss, the lovasz loss\cite{lovasz} and the wavelet regularization terms. These components are structured as:
\begin{equation}
\mathcal{L}=\mathcal{L}_{\mathrm{ce}}+ \mathcal{L}_{\text {lovasz}} + \mathcal{L}_{\mathrm{wr}}
\end{equation}
\begin{equation}
\mathcal{L}_{\mathrm{wr}} = \lambda_{1} \sum_{t=1}^{N}\left\|\mathcal{D}_{t}\right\|_{2}^{2}+\lambda_{2} \sum_{t=1}^{N}\left\|\mathcal{A}_{t}-\mathcal{A}_{t-1}\right\|_{2}^{2} ,
\end{equation}
where $\mathcal{L}_{\mathrm{ce}}$ represents the cross-entropy loss, assessing the accuracy of noise predictions, and $\mathcal{L}_{\text{lovasz}}$ denotes the lovasz loss, which is designed for improving performance in tasks with imbalanced classes. $\mathcal{L}_{\mathrm{wr}}$ represents the wavelet regularization terms, which include two term. The first wavelet regularization term minimizes the sum of the $l_{2}$ norm of the detail components across all levels, enhancing detail preservation. The second wavelet term minimizes the $l_{2}$ norm of the difference between the approximation components of consecutive levels across all levels, crucial for preserving the mean of the input signal and ensuring a proper wavelet decomposition
enforcing a consistent wavelet decomposition structure, which is essential for effectively capturing multi-scale geometric information. $\lambda_{1}$ and $\lambda_{2}$ are the wavelet regularization parameters used to adjust the strength of the regularization terms. $\mathcal{D}_{t}$ represents the mean of the detail sub-bands at decomposition level $t$, while $\mathcal{A}_{t}$ and $\mathcal{A}_{t-1}$ denote the mean of the approximation sub-bands at levels $t$ and level ${t-1}$ of the lifting wavelet, respectively.

\section{Denoising and Perception Benchmarks under Adverse Weather}

In this section, we establish a Denoising benchmark to evaluate the performance of our denoising model and introduce three downstream LiDAR perception benchmarks: Semantic Segmentation (SS), Place Recognition (PR), and Object Detection (OD), to assess the generalization of state‑of‑the‑art perception models under adverse weather and the effectiveness of our denoising model as a preprocessing step. Notably, in all downstream benchmarks, our denoising model is trained in a supervised manner solely on our Weather‑KITTI and Weather‑NuScenes datasets using only point‑wise weather labels. Meanwhile, all perception models are directly tested on real‑world adverse‑weather datasets without any retraining or fine‑tuning. This design highlights the strong transferability of our datasets and the practical value of the proposed denoising models for real‑world applications. All experiments were conducted on an Ubuntu 18.04 with Intel Xeon Platinum 8280 CPUs, two Nvidia V100 GPUs, and 32 GB RAM.




\begin{table}[t]
\captionsetup{justification=centering, singlelinecheck=false}
\caption{Splitting of Training, Validation, and Testing Sets for Our Datasets.}
\label{weather_simulations}
\centering
\renewcommand{\arraystretch}{0.9} 
\begin{tabular}{c@{\hskip 0.2cm}c@{\hskip 0.2cm}p{2cm}@{\hskip 0.2cm}p{3cm}}
\toprule
& & {Weather-KITTI} & {Weather-NuScenes} \\
\midrule
\addlinespace[2pt]
\multirow{5}{*}{\textbf{Snow Scenarios}} 
& {Train:} &{\raggedright [01,02,08,17,19]} 
& {\raggedright [01,02,07,08]} \\
\cline{2-4}
\addlinespace[2pt]

& {Val:} & {\raggedright [04,11,12,16]} 
& {\raggedright [04]} \\
\cline{2-4}
\addlinespace[2pt]

& \multirow{3}{*}{Test:} & {\raggedright [00,03,05,06,07,\\
09,10,13,14,15,\\
18,20,21]} 
& \multirow{3}{*}{\raggedright [00,03,05,06,09]} \\

\midrule

\multirow{5}{*}{\textbf{Fog Scenarios}} 
& \multirow{2}{*}{Train:} & {\raggedright [02,03,04,16,17,\\
18,19,20]} 
& \multirow{2}{*}{\raggedright [01,02,03,07]} \\
\cline{2-4}
\addlinespace[2pt]

& {Val:} & {\raggedright [06,07,09,11]} 
& {\raggedright [04]} \\
\cline{2-4}
\addlinespace[2pt]

& \multirow{2}{*}{Test:} & {\raggedright [00,01,05,08,10,\\
12,13,14,15,21]} 
& \multirow{2}{*}{\raggedright [00,05,06,08,09]} \\
\midrule

\multirow{5}{*}{\textbf{Rain Scenarios}} 
& \multirow{2}{*}{Train:} & {\raggedright [01,02,03,04,06,\\
12,16,17,21]} 
& \multirow{2}{*}{\raggedright [04,05,07,09]} \\
\cline{2-4}
\addlinespace[2pt]

& {Val:} & {\raggedright [09,19,19]} 
& {\raggedright [08]} \\
\cline{2-4}
\addlinespace[2pt]

& \multirow{2}{*}{Test:} & {\raggedright [00,05,07,08,10,\\
13,14,15,18,20]} 
& \multirow{2}{*}{\raggedright [00,01,02,03,06]} \\
\bottomrule
\end{tabular}
\label{setdivision}
\vspace{-1.2em}
\end{table}

\footnotetext[1]{The training set is: [14,15,18,20,24,28,34,36,37].
The val set is: [11,16].
The test set is: [12,13,17,22,23,26,30,35,76].}

\subsection{ Denoising Benchmark}
\textbf{Benchmark Setting:} We conduct denoising experiments on the real‑world extreme‑weather dataset WADS~\cite{dsor} and our Weather‑KITTI and Weather‑NuScenes datasets. WADS\cite{dsor} comprises 1,300 LiDAR scans across 20 sequences, collected in snowy urban driving scenarios. We divided the WADS dataset\footnotemark[1] into training, validation, and testing sets according to the protocol outlined in\cite{3DOutDet}. To ensure a fair comparison, we re-trained state-of-the-art models using our splits. For Weather‑KITTI and Weather‑NuScenes, the dataset partitions for training, validation, and testing are provided in Table~\ref{setdivision}. During training, point clouds were downsampled to a 10 cm voxel grid and projected onto the X, Y, and Z planes at [256, 256, 32] resolutions. Models were trained for 30 epochs using AdamW~\cite{adam} with a weight decay of 0.005, a batch size of 4 per GPU, and stochastic depth (drop probability 0.2). The learning rate warmed up linearly to 0.001 over 2 epochs, then decayed to $10^{-5}$ with cosine annealing. The loss combined cross‑entropy, Lovasz, and wavelet regularization with $\lambda_1=\lambda_2=0.1$.

Following prior work~\cite{3DOutDet,4DenoiseNet,wetahernet}, we evaluate the point cloud denoising task using precision, recall, F1 score, and mean Intersection‑over‑Union (mIoU) as quantitative metrics, where $precision=\frac{TP}{TP+FP}$ and $recall=\frac{TP}{TP+FN}$. Specifically, the F1 score and mIoU are defined as
\begin{equation}	
	{F_{1}} = 2 \times \frac{precision \times recall}{{precision + recall}} \\
\end{equation}
\begin{equation}	
	{mIoU} = \frac{TP}{{TP+FP+FN}}
\end{equation}
where $TP$ denotes correctly identified noise points, $FP$ denotes non‑noise points incorrectly identified as noise, and $FN$ denotes noise points that were not correctly identified.

\textbf{Results On WADS:} 
We compare our model with the current state‑of‑the‑art models, with a primary focus on point cloud denoising networks. Specifically, WeatherNet~\cite{wetahernet}, 4DenoiseNet~\cite{4DenoiseNet}, and 3D‑OutDet~\cite{3DOutDet} are specialized point cloud denoising networks. In addition, SOR~\cite{sor}, ROR~\cite{sor}, DSOR~\cite{dsor}, and DROR~\cite{dror} are statistical methods, while SalsaNext~\cite{Salsanext}, Cylinder3D~\cite{cylinder3d}, and LSK3DNet~\cite{feng2024lsk3dnet} are point cloud segmentation methods that we have adapted for binary classification to enable comparison. Table \ref{wads} presents a comprehensive comparison of denoising performance on the WADS\cite{dsor} dataset, evaluating precision, recall, F1 score, and mIoU. As illustrated in Table \ref{wads}, while LSK3DNet\cite{feng2024lsk3dnet} achieves the highest precision with a score of 97.16, our model surpasses all existing methods in other critical metrics, achieving the highest recall, F1 score, and mIoU with scores of 93.93, 95.13, and 90.73, respectively. These results underscore the superior efficacy of our model in point cloud denoising, particularly in complex real-world applications.

\begin{table}[t]
\captionsetup{justification=centering, singlelinecheck=false}
\centering
\caption{{Denoising Performance Results on WADS Datasets.}}
\begin{tabular}{lcccccc}
\toprule
\textbf{Method} & \textbf{Precision↑} & \textbf{Recall↑} & \textbf{F1↑} & \textbf{mIoU↑}  \\
\midrule
SalsaNext\cite{Salsanext} & 74.16 & 93.50 & 82.71 & 70.52   \\
Cylinder3D\cite{cylinder3d} & {97.12} & 92.61 & 94.81 & 90.13  \\
LSK3DNet\cite{feng2024lsk3dnet} &\textbf{97.16}  &92.70  &94.88  &90.25   \\
\midrule
ROR\cite{sor} & 17.13 & 91.80 & 29.15 & 17.06 \\
DSOR\cite{dsor} & 65.92 & 90.93 & 76.43 & 61.86   \\
DROR\cite{dror} & 69.84 & 90.10 & 78.68 & 64.05  \\
WeatherNet\cite{wetahernet} & {96.69} & 81.24 & 88.28 & 79.02   \\
4DenoiseNet\cite{4DenoiseNet} & 96.46 & 86.01 & 90.94 & 83.38  \\
3D-OutDet\cite{3DOutDet} & 97.10 & 92.25 & 94.61 & 89.78  \\
\hline
\textbf{TripleMixer (Ours)} & 96.38 & \textbf{93.93} & \textbf{95.13} & \textbf{90.73}   \\
\bottomrule
\label{wads}
\end{tabular}
\vspace{-1.2em}
\end{table}

\begin{figure*}[t]
	\centerline{\includegraphics[width=\linewidth]{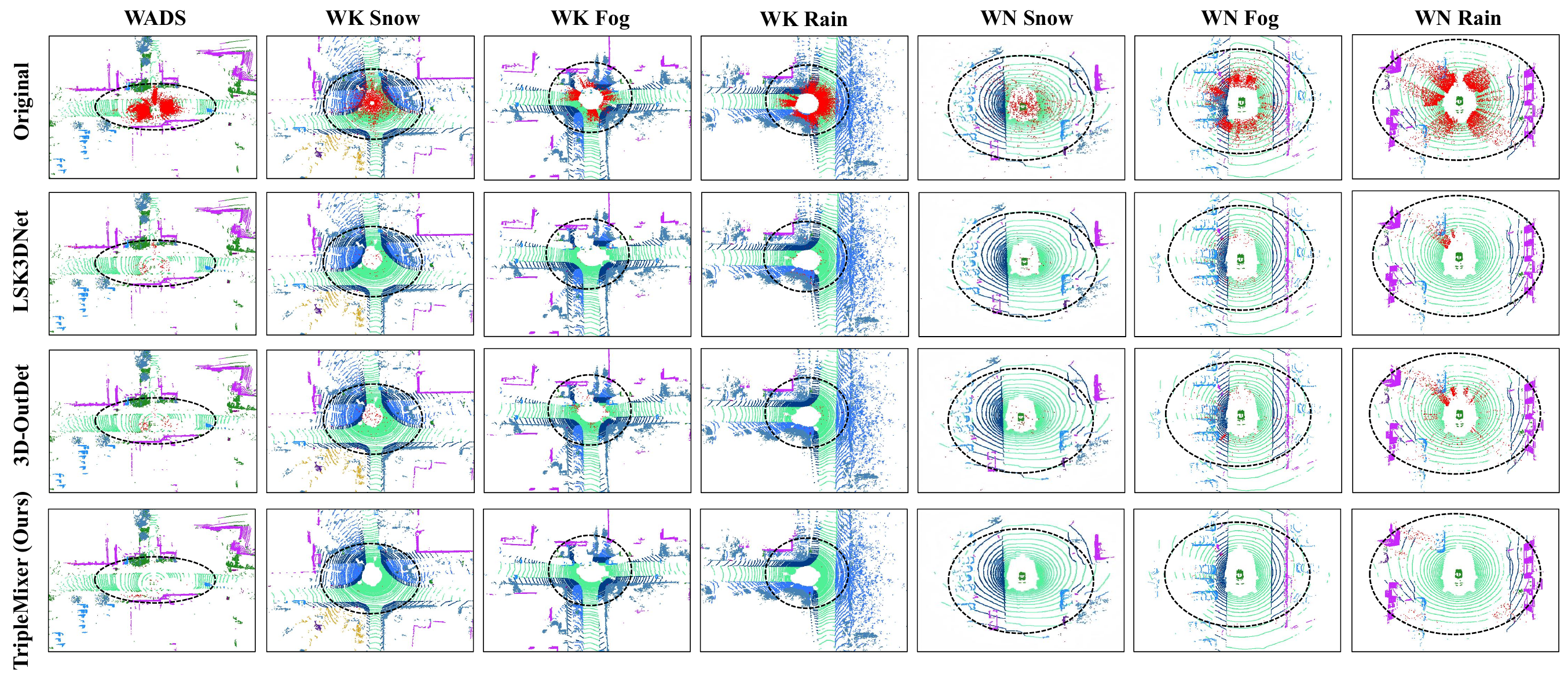}}
	\caption{ Qualitative results of the Denoising benchmark on the WADS~\cite{dsor}, Weather‑KITTI (WK), and Weather‑NuScenes (WN) datasets.} Semantic point colors follow the scheme in Figure~\ref{distrubution}, with \textcolor{red}{red points} indicating weather‑induced noise.
	\label{denoise-vis}

\end{figure*}

\begin{table*}[h]
\centering
\caption{\centering{Denoising Performance On The Proposed Weather-KITTI Datasets. Average mIoU Represents the Mean mIOU Across Three Weather Scenarios.}}
\begin{tabular}{@{}l@{\hspace{1.0em}}c@{\hspace{1.0em}}c@{\hspace{1.0em}}c@{\hspace{1.0em}}c@{\hspace{1.0em}}c@{\hspace{1.0em}}c@{\hspace{1.0em}}c@{\hspace{1.0em}}c@{\hspace{1.0em}}c@{\hspace{1.0em}}c@{\hspace{1.0em}}c@{\hspace{1.0em}}c@{\hspace{1.0em}}c@{}}
\toprule
\multirow{2}{*}{\textbf{Method}} &\textbf{Average}  & \multicolumn{4}{c}{\textbf{Snow Scenarios}} & \multicolumn{4}{c}{\textbf{Fog Scenarios}} & \multicolumn{4}{c}{\textbf{Rain Scenarios}} \\

\cmidrule(lr){3-6} \cmidrule(lr){7-10} \cmidrule(lr){11-14}
&\textbf{mIoU}  & \textbf{Precision↑} & \textbf{Recall↑} & \textbf{F1↑} & \textbf{mIoU↑} 
& \textbf{Precision↑} & \textbf{Recall↑} & \textbf{F1↑} & \textbf{mIoU↑} 
& \textbf{Precision↑} & \textbf{Recall↑} & \textbf{F1↑} & \textbf{mIoU↑} \\
\midrule
SalsaNext\cite{Salsanext} &89.58 &97.40 &92.83 &95.06 &90.58 &97.19 &94.59 &95.87 &92.07 &\textbf{97.75} &87.82 &92.52 &86.08 \\
Cylinder3D\cite{cylinder3d} &94.28 &99.13 &96.39 &97.74 &95.58 &\textbf{99.78} &92.92 &96.23 &92.73 &96.79 &97.58 &97.18 &94.52 \\
LSK3DNet\cite{feng2024lsk3dnet} &94.66 &98.95 &96.85 &97.87 &95.94 &99.60 &93.75 &96.51 &93.16 &97.20 &97.62 &97.40 & 94.88 \\
\midrule
DROR\cite{dror} &56.24 &44.18 &\textbf{97.36} &60.77 &43.66 &66.05 &92.74 &78.63 &64.79 &61.47 &96.68 &74.11 &60.28 \\
WeatherNet\cite{wetahernet} &79.15 &83.28 &{97.20} &89.70 &81.33 &89.16 &86.64 &{87.88} &78.38 &78.49 &98.81 &87.48 &77.75 \\ 
4DenoiseNet\cite{4DenoiseNet} &81.85 &85.23 &97.08 &90.77 &83.10 &89.90 &93.95 &91.88 &81.98 &83.02 &96.34 &89.18 &80.48 \\
3D-OutDet\cite{3DOutDet} &{91.55} &98.14 &95.15 &96.63 &93.48 &{99.61} &85.26 &91.88 &87.97 &94.12 &98.98 &96.48 &93.21 \\
\hline
\textbf{TripleMixer (Ours)} &\textbf{96.31} &\textbf{99.23} &97.18 &\textbf{98.16} &\textbf{96.46} &99.03 &\textbf{96.50} &\textbf{97.74} &\textbf{95.61} &{97.66} &\textbf{99.15} &\textbf{98.40} &\textbf{96.87} \\
\bottomrule
\label{Weather-KITTI}
\end{tabular}
\vspace{-1.2em}
\end{table*}

\begin{table*}[h]
\centering
\caption{\centering{Denoising Performance Results On The Proposed Weather-NuScenes Datasets. Average mIoU Represents the Mean mIOU Across Three Weather Scenarios: Snow, Fog, and Rain.}}
\begin{tabular}{@{}l@{\hspace{1.0em}}c@{\hspace{1.0em}}c@{\hspace{1.0em}}c@{\hspace{1.0em}}c@{\hspace{1.0em}}c@{\hspace{1.0em}}c@{\hspace{1.0em}}c@{\hspace{1.0em}}c@{\hspace{1.0em}}c@{\hspace{1.0em}}c@{\hspace{1.0em}}c@{\hspace{1.0em}}c@{\hspace{1.0em}}c@{}}
\toprule
\multirow{2}{*}{\textbf{Method}} &\textbf{Average}  & \multicolumn{4}{c}{\textbf{Snow Scenarios}} & \multicolumn{4}{c}{\textbf{Fog Scenarios}} & \multicolumn{4}{c}{\textbf{Rain Scenarios}} \\

\cmidrule(lr){3-6} \cmidrule(lr){7-10} \cmidrule(lr){11-14}
&\textbf{mIoU}  & \textbf{Precision↑} & \textbf{Recall↑} & \textbf{F1↑} & \textbf{mIoU↑} 
& \textbf{Precision↑} & \textbf{Recall↑} & \textbf{F1↑} & \textbf{mIoU↑} 
& \textbf{Precision↑} & \textbf{Recall↑} & \textbf{F1↑} & \textbf{mIoU↑} \\
\midrule
SalsaNext\cite{Salsanext} &73.12 &80.48 &75.23 &77.76 &63.21 &93.79 &84.21 &88.74 &79.76 &82.74 &90.89 &86.62 &76.40 \\
Cylinder3D\cite{cylinder3d} &95.41 &\textbf{98.85} &95.56 &97.18 &94.51 &98.94 &96.37 &97.64 &95.38 &99.24 &97.06 &98.14 &96.34 \\
LSK3DNet\cite{feng2024lsk3dnet} &96.04 &98.55 &97.20 &97.86 &94.96 &99.10 &97.70 &98.39 &96.08 &99.35 &98.00 &98.66 &97.10 \\
\midrule
DROR\cite{dror} &42.93 &38.36 &92.74 &54.27 &37.24 &40.36 &\textbf{98.68} &57.28 &40.14 &51.71 &98.89 &67.91 &51.41 \\
WeatherNet\cite{wetahernet} &71.73 &72.96 &85.40 &78.69 &64.87 &84.97 &80.54 &82.69 &70.49 &80.44 &\textbf{99.06} &88.78 &79.83 \\
4DenoiseNet\cite{4DenoiseNet} &84.38 &78.97 &96.86 &87.01 &76.99 &92.34 &97.53 &94.86 &90.23 &87.56 &97.88 &92.43 &85.93 \\
3D-OutDet\cite{3DOutDet} &90.87 &95.17 &90.43 &92.74 &86.46 &97.53 &90.16 &93.70 &88.15 &99.44 &98.53 &98.98 &97.99  \\
\hline
\textbf{TripleMixer (Ours)} &\textbf{97.21} &97.32 &\textbf{98.37} &\textbf{97.84} &\textbf{95.78} &\textbf{99.37} &{98.02} &\textbf{98.69} &\textbf{97.42} &\textbf{99.72} &{98.71} &\textbf{99.21} &\textbf{98.43} \\
\bottomrule
\label{Weather-NuScenes}
\vspace{-1.2em}
\end{tabular}
\end{table*}

\textbf{Results On Weather-KITTI:} To further validate the good generalizability of our model, we conducted additional denoising comparison experiments on our proposed Weather-KITTI and Weather-NuScenes datasets. For a thorough evaluation, we selected the most recent and best-performing models from Table \ref{wads} for further comparative analysis. As shown in Table \ref{Weather-KITTI}, we compared our model with seven state-of-the-art methods across three scenarios in the Weather-KITTI dataset: Snow Scenarios, Fog Scenarios and Rain Scenarios. Our model achieved the highest average mIoU of 96.31, outperforming general semantic segmentation methods such as LSK3DNet~\cite{feng2024lsk3dnet} and advanced denoising models like 4DenoiseNet~\cite{4DenoiseNet} and 3D‑OutDet~\cite{3DOutDet}. Specifically, our model's average mIoU exceeded LSK3DNet, 4DenoiseNet, and 3D‑OutDet by 1.7\%, 16.2\%, and 5.9\%, respectively. Notably, our model achieved the highest mIoUs of 96.46, 95.61, and 96.87 in Snow, Fog, and Rain scenarios, respectively, whereas Cylinder3D~\cite{cylinder3d}, and SalsaNext~\cite{Salsanext} attained the highest Precision in individual scenarios. This likely reflects their conservative filtering or task‑specific feature modeling, which improves precision by suppressing false positives but often reduces recall. In contrast, our Frequency Mixer (FMX) module performs multi‑scale frequency decomposition through a learnable lifting‑wavelet design, enabling the selective retention of task‑relevant high‑frequency details. This design prioritizes structural completeness over aggressive filtering, thereby striking a better balance between precision and recall and ensuring robust and consistent denoising across diverse weather conditions.

\textbf{Results On Weather-NuScenes:} We also compared the denoising performance of our model with seven state-of-the-art methods across three scenarios in the Weather-NuScenes dataset. As shown in Table \ref{Weather-NuScenes}, our TripleMixer model achieved the highest average mIoU of 97.21 across these three scenarios. Notably, the average mIoU of our model is 6.9\% higher than 3D-OutDet\cite{3DOutDet} and 15.20\% higher than 4DenoiseNet\cite{4DenoiseNet}. Additionally, our model achieved the best mIoU accuracy in each scenario (Snow, Fog, and Rain) compared to other denoising models, with mIoUs of 95.78, 97.42, and 98.48, respectively. These results demonstrate that our model exhibits superior performance in the Weather-NuScenes dataset across three different adverse weather scenarios compared to other denoising models. 

Figure~\ref{denoise-vis} visualizes denoising results on WADS~\cite{dsor}, Weather‑KITTI, and Weather‑NuScenes, showing that TripleMixer consistently produces more accurate and cleaner reconstructions than 3D‑OutDet~\cite{3DOutDet} and LSK3DNet~\cite{feng2024lsk3dnet} across various weather conditions.


\begin{table*}[t] 
\caption{\centering{Benchmarking Results for the SS Task (\%). All SS models are pretrained on SemanticKITTI~\cite{Semantickitti} and inferred on SemanticSTF~\cite{xiao20233d}. Denoising models are trained on Weather-KITTI and used as preprocessing during inference.}}
\centering
\footnotesize

\setlength{\tabcolsep}{1.6pt} 
\renewcommand\arraystretch{1.1}

\begin{tabular}{lcc cccc cccc cccc cccc ccc}
\toprule
\multicolumn{2}{c}{\textbf{Methods}} &
\multirow{2}{*}{\textbf{mIoU}} &
\multirow{2}{*}{\rotatebox{90}{\textbf{car}}} &
\multirow{2}{*}{\rotatebox{90}{\textbf{bi.cle}}} &
\multirow{2}{*}{\rotatebox{90}{\textbf{mt.cle}}} &
\multirow{2}{*}{\rotatebox{90}{\textbf{truck}}} &
\multirow{2}{*}{\rotatebox{90}{\textbf{oth-v.}}} &
\multirow{2}{*}{\rotatebox{90}{\textbf{pers.}}} &
\multirow{2}{*}{\rotatebox{90}{\textbf{bi.list}}} &
\multirow{2}{*}{\rotatebox{90}{\textbf{mt.list}}} &
\multirow{2}{*}{\rotatebox{90}{\textbf{road}}} &
\multirow{2}{*}{\rotatebox{90}{\textbf{parki.}}} &
\multirow{2}{*}{\rotatebox{90}{\textbf{sidew.}}} &
\multirow{2}{*}{\rotatebox{90}{\textbf{oth-g.}}} &
\multirow{2}{*}{\rotatebox{90}{\textbf{build.}}} &
\multirow{2}{*}{\rotatebox{90}{\textbf{fence}}} &
\multirow{2}{*}{\rotatebox{90}{\textbf{veget.}}} &
\multirow{2}{*}{\rotatebox{90}{\textbf{trunk}}} &
\multirow{2}{*}{\rotatebox{90}{\textbf{terra.}}} &
\multirow{2}{*}{\rotatebox{90}{\textbf{pole.}}} &
\multirow{2}{*}{\rotatebox{90}{\textbf{traf.}}} 
\\
\cmidrule(lr){1-2}
\textbf{SS} & \textbf{Preprocessing} & & & & & & & & & & & & & & & & & & & & \\
\midrule

\multirow{4}{*}{{SphereFormer~\cite{lai2023spherical}}} & Baseline  &11.26 &19.12 &0.00 &0.00 &13.83 &\textbf{4.12} &0.99 &\textbf{6.75} &0.00 &22.28 &4.69 &15.80 &4.43 &\textbf{25.53} &13.27 &24.23 &4.38 &37.74 &14.71 &1.99
\\
&4DenoiseNet~\cite{4DenoiseNet}		&11.03 &18.24 &0.00 &0.00 &13.42 &3.92 &1.00 &2.17 &0.00 &22.01 &4.72 &16.25 &4.74 &25.33 &14.68 &23.97 &4.50 &38.15 &14.57 &1.98 \\
&3D-OutDet~\cite{3DOutDet}	&11.87 &23.26 &0.00 &0.00 &15.94 &3.58 &\textbf{1.64} &1.54 &0.00 &24.44 &4.68 &17.41 &5.73 &25.04 &13.99 &\textbf{27.25} &\textbf{5.30} &38.57 &\textbf{15.20} &\textbf{2.04} \\
&{TripleMixer (Ours)}	&\textbf{13.71} &\textbf{27.19} &0.00 &0.00 &\textbf{21.20} &4.07 &0.21 &5.68 &0.00 &\textbf{26.72} &\textbf{4.76} &\textbf{19.68} &\textbf{8.85} &23.97 &\textbf{33.64} &22.86 &4.31 &\textbf{40.95} &14.48 &1.84\\
\midrule

\multirow{4}{*}{{SFPNet \cite{wang2024sfpnet}}} & Baseline  &11.14 &14.36 &0.00 &0.00 &11.60 &0.00 &0.23 &\textbf{5.29} &0.00 &38.06 &4.33 &13.71 &5.09 &17.19 &12.06 &30.01 &4.97 &37.98 &14.86 &1.92
\\
&4DenoiseNet~\cite{4DenoiseNet}		&11.07 &14.01 &0.00 &0.00 &10.89 &0.00 &0.24 &4.55 &0.00 &38.18 &4.12 &13.73 &5.08 &17.26 &12.39 &29.73 &4.96 &38.07 &15.10 &1.95 \\
&3D-OutDet~\cite{3DOutDet}	&12.12 &\textbf{23.51} &0.00 &0.00 &11.27 &0.00 &0.34 &4.14 &0.00 &\textbf{43.87} &6.45 &14.83 &5.81 &17.30 &11.76 &30.38 &4.88 &\textbf{38.37} &\textbf{15.42} &\textbf{2.01} \\
&{TripleMixer (Ours)}	&\textbf{13.23} &21.33 &\textbf{0.06} &0.00 &\textbf{15.95} &0.00 &\textbf{0.36} &4.91 &0.00 &42.35 &\textbf{8.37} &\textbf{15.74} &\textbf{6.58} &\textbf{19.88} &\textbf{18.61} &\textbf{37.01} &\textbf{5.05} &38.00 &15.39 &1.87 \\

\midrule

\multirow{4}{*}{{PTv3~\cite{wu2024point}}} & Baseline  &14.58 &15.27 &1.13 &0.00 &3.56 &0.40 &0.21 &0.00 &37.25 &44.10 &\textbf{12.05} &\textbf{9.95} &6.46 &25.30 &25.70 &37.46 &8.63 &36.97 &10.10 &2.46
\\
&4DenoiseNet~\cite{4DenoiseNet}		&14.53 &14.88 &1.12 &0.00 &3.60 &0.48 &0.21 &0.00 &37.25 &44.07 &11.86 &9.92 &6.43 &25.42 &25.95 &37.57 &8.70 &37.00 &9.22 &2.48 
\\
&3D-OutDet~\cite{3DOutDet}	&16.40 &19.87 &2.14 &0.00 &9.58 &2.48 &0.21 &0.00 &{40.56} &\textbf{44.51} &11.30 &9.87 &8.48 &25.28 &\textbf{30.03} &\textbf{40.60} &8.54 &\textbf{42.18} &\textbf{12.12} &\textbf{2.79} 
\\
&{TripleMixer (Ours)}	&\textbf{17.18} &\textbf{23.64} &\textbf{2.80} &0.00 &\textbf{19.88} &\textbf{8.26} &\textbf{1.74} &0.00 &\textbf{42.87} &44.06 &11.36 &9.91 &\textbf{9.36} &\textbf{29.09} &28.53 &36.93 &\textbf{9.26} &36.93 &9.87 &1.98 
\\
\bottomrule
\label{LSS_BENCH}
\end{tabular}
\vspace{-1.0em}
\end{table*}

\begin{figure}[!t]
\centerline{\includegraphics[width=\linewidth]{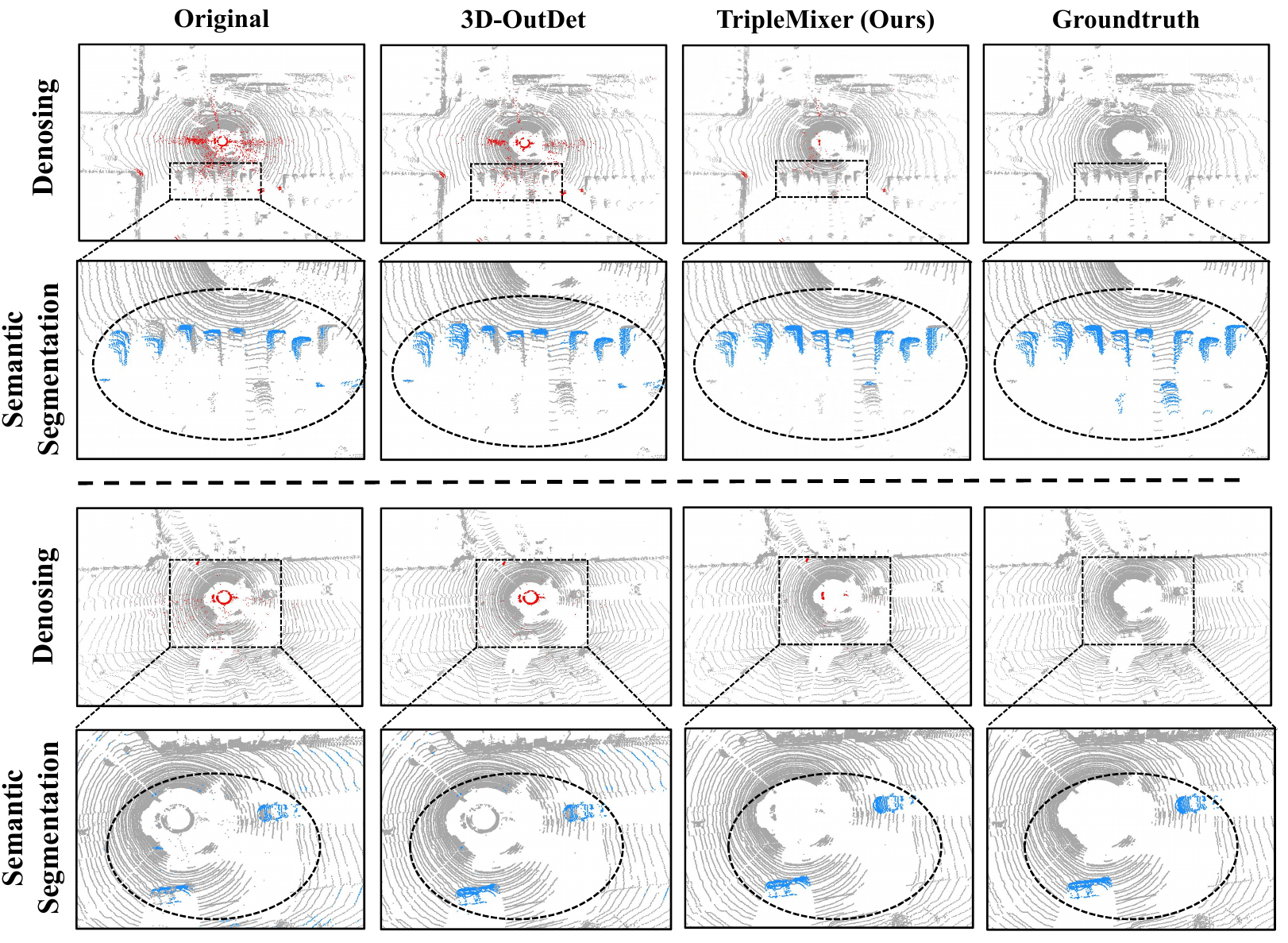}}
	\caption{Qualitative results of the SS benchmark under adverse weather. The “Denoising” row presents outputs of different denoising models, while the “Semantic Segmentation” row shows PTv3\cite{wu2024point} results on the corresponding denoised point clouds. \textcolor{red}{Red points} denote weather‑induced noise, and \textcolor{semanticblue}{blue points} represent the “car” class.}
	\label{LSS-vis}
 \vspace{-0.9em}
\end{figure}

\subsection{ Semantic Segmentation (SS) Benchmark} 
\textbf{Benchmark Setting:}
We evaluate the downstream SS models on the real‑world adverse‑weather SemanticSTF~\cite{xiao20233d} dataset, which provides semantic annotations for 19 categories consistent with SemanticKITTI. To comprehensively assess performance, we select three recent state‑of‑the‑art semantic segmentation models: SphereFormer~\cite{lai2023spherical}, SFPNet~\cite{wang2024sfpnet}, and Point Transformer v3 (PTv3)~\cite{wu2024point}. In our experiments, they pretrained on SemanticKITTI under clear‑weather conditions are directly evaluated on SemanticSTF. Furthermore, to examine the impact of denoising on SS models performance, we incorporate three specialized denoising networks: 4DenoiseNet\cite{4DenoiseNet}, 3D‑OutDet~\cite{3DOutDet}, and our proposed TripleMixer, all trained in a supervised manner on the our Weather‑KITTI datasets and directly generalized to SemanticSTF. Evaluation follows standard protocols, computing the Intersection‑over‑Union (IoU) for each class and reporting the mean IoU (mIoU) across all classes.

\textbf{Benchmark Results:}
Table~\ref{LSS_BENCH} presents the benchmarking results for the SS task under real-world adverse weather on the validation set of SemanticSTF~\cite{xiao20233d} dataset. The baseline denotes direct application of SS models without denoising preprocessing. Across all segmentation backbones, including SphereFormer~\cite{lai2023spherical}, SFPNet~\cite{wang2024sfpnet}, and PTv3~\cite{wu2024point}, the proposed TripleMixer consistently yields the highest mIoU, demonstrating significant improvements over both the baseline and other denoising methods. Specifically, when integrated with SphereFormer, TripleMixer achieves an mIoU of 13.71, improving by 2.45 points over the baseline. Similar gains are observed with SFPNet and PTv3, where TripleMixer improves the mIoU by 2.09 and 2.60 points, respectively. On average, TripleMixer achieves a 2.38-point improvement across all three backbones, significantly surpassing the performance of 3D-OutDet, which only provides an average gain of 1.13 points. Notably, 4DenoiseNet, which demonstrates limited denoising effectiveness on our proposed datasets, fails to improve real-world segmentation performance and even leads to performance degradation compared to the baseline. At the category level, TripleMixer delivers substantial gains in key semantic classes, improving IoU by 7.81 for car and 9.35 for truck over the baseline, averaged across the three segmentation models. It also enhances performance in challenging structural categories such as buildings and fences. These results underscore the robustness and practical utility of TripleMixer and our dataset for improving LiDAR semantic segmentation in real‑world conditions. We visualize the SS benchmark under adverse weather in Figure~\ref{LSS-vis} to illustrate denoising results in real‑world scenarios and the subsequent semantic segmentation performance, focusing on the key semantic class “car”. Compared with 3D‑OutDet, TripleMixer achieves more effective denoising and enhances segmentation performance under real‑world extreme conditions.

\begin{table*}[h]
\centering
\footnotesize
\captionsetup{aboveskip=2pt, belowskip=0pt}
\renewcommand\arraystretch{1.1}
\setlength{\tabcolsep}{0.6pt}
\caption{\centering{Benchmarking Results for the PR Task. All PR models are trained on the KITTI~\cite{kitti} and inferred on Boreas (BR)~\cite{Boreas} and Weather-KITTI (WK). Denoising models, trained on Weather-KITTI, serve as preprocessing during inference.}}
\begin{tabular}{lcc@{\hspace{8pt}}ccc@{\hspace{8pt}}ccc@{\hspace{8pt}}ccc@{\hspace{8pt}}ccc@{\hspace{8pt}}ccc}
\toprule
\multicolumn{2}{c}{\textbf{Method}} &\multirow{2}{*}{\textbf{$\text{mRS}$↑}}  & \multicolumn{3}{c}{\textbf{WK Snow}} & \multicolumn{3}{c}{\textbf{WK Fog}} & \multicolumn{3}{c}{\textbf{WK Rain}} & \multicolumn{3}{c}{\textbf{BR Easy}} & \multicolumn{3}{c}{\textbf{BR Hard}}\\
\cmidrule(lr){1-2} \cmidrule(lr){4-6} \cmidrule(lr){7-9} \cmidrule(lr){10-12} \cmidrule(lr){13-15} \cmidrule(lr){16-18}
\textbf{PR}  &\textbf{Preprocessing}  & 
& \textbf{R@1↑}  & \textbf{R@1\%↑}   & \textbf{F1↑}
& \textbf{R@1↑}  & \textbf{R@1\%↑}   & \textbf{F1↑}
& \textbf{R@1↑}  & \textbf{R@1\%↑}   & \textbf{F1↑}
& \textbf{R@1↑}  & \textbf{R@1\%↑}   & \textbf{F1↑}
& \textbf{R@1↑}  & \textbf{R@1\%↑}   & \textbf{F1↑}\\
\midrule
\multirow{4}{*}{{OT \cite{ma2022overlaptransformer}}} &Baseline &0.15 &0.03 &0.26	&0.07		&0.09 &0.33	&0.17	&0.17	&0.52 	&0.29  &0.01	&0.11 	&0.02 &0.00 &0.09 &0.02
\\
&4DenoiseNet~\cite{4DenoiseNet}	&0.22 &0.03	&0.24	&0.06	&0.21 &0.58	&0.37	&0.25	&0.61 	&0.40  &0.01	&0.09 	&0.02 &0.00 &0.06 &0.01
\\
&3D-OutDet~\cite{3DOutDet}	&0.27 &0.07 &0.38	&0.13	&0.22 &0.69	&0.42	&0.27	&0.62 	&0.43 &0.02	&0.13 	&0.03 &0.01 &0.11 &0.03
\\
&TripleMixer (Ours)	&\textbf{0.31} &\textbf{0.09} &\textbf{0.43}	&\textbf{0.17}	&\textbf{0.31}	&\textbf{0.78} &\textbf{0.47}	&\textbf{0.29}	&\textbf{0.65}	&\textbf{0.44} 	&\textbf{0.04} &\textbf{0.15}	&\textbf{0.06} 	&\textbf{0.03} &\textbf{0.14} &\textbf{0.05}
\\

\midrule 
\multirow{4}{*}{{CVTNet~\cite{ma2023cvtnet}	}} &Baseline &0.28 &0.03	&0.34	&0.07	&0.29  &0.50	&0.46 &0.19	&0.61 	&0.33  &0.14	&0.21 	&0.25 &0.09 &0.14 &0.15
\\
&4DenoiseNet~\cite{4DenoiseNet}	&0.33 &0.02	&0.30	&0.03	&0.31 &0.48	&0.46	&0.43	&0.83	&0.60  &0.14	&0.20 	&0.25 &0.10 &0.15 &0.17
\\
&3D-OutDet~\cite{3DOutDet}	&0.37 &0.04	&0.42	&0.09	&0.31 &0.50	&0.47	&0.49	&0.86 	&0.65 &\textbf{0.17}	&0.23 	&0.26 &0.12 &0.17 &0.19
\\
&TripleMixer (Ours)		&\textbf{0.39} &\textbf{0.06}	&\textbf{0.44}	&\textbf{0.11}	&\textbf{0.33} &\textbf{0.54}	&\textbf{0.50}	&\textbf{0.51}	&\textbf{0.89} 	&\textbf{0.67}  &{0.16}	&\textbf{0.29} 	&\textbf{0.28} &\textbf{0.13} &\textbf{0.24} &\textbf{0.21}
\\
\midrule 

\multirow{4}{*}{{LPSNet~\cite{liu2024lps}	}} &Baseline &0.30 &0.02	&0.18	&0.04	&0.09 &0.77	&0.16	&0.29	&0.74	&0.46  &0.13	&0.28 	&0.26 &0.10 &0.26 &0.19
\\
&4DenoiseNet~\cite{4DenoiseNet}	&0.29 &0.02	&0.22	&0.05	&0.03 &0.77	&0.06	&0.28	&0.72 	&0.44  &0.16 &0.28 &0.27 &0.10	&0.24 	&0.19
\\
&3D-OutDet~\cite{3DOutDet}	&0.33 &0.02	&0.27	&\textbf{0.06}	&0.10 &\textbf{0.85}	&{0.17}	&0.32	&0.75 	&0.49  &\textbf{0.18}	&0.29 	&0.28 &\textbf{0.11} &0.27 &0.20
\\
&TripleMixer (Ours)		&\textbf{0.37}  &\textbf{0.03}	&\textbf{0.29}	&\textbf{0.06}	&\textbf{0.14} &0.83	&\textbf{0.24}	&\textbf{0.37}	&\textbf{0.79} 	&\textbf{0.54} &{0.15} &\textbf{0.36} 	&\textbf{0.29} &\textbf{0.11} &\textbf{0.34} &\textbf{0.23} 
\\

\bottomrule
\label{LPR_BENCH}
\end{tabular}
\vspace{-0.3cm}
\end{table*}

\subsection{ Place Recognition (PR) Benchmark} 
\textbf{Benchmark Setting:}
We selected distinct sequences from the real‑world Boreas~\cite{Boreas} dataset and our Weather‑KITTI dataset. In Boreas, the sequence recorded on 2021-11-02-11-16 is used as the database, while the sequence from 2021-01-26-11-22, which features heavy snowfall, serves as the query. Boreas Easy employs only the forward traversal, whereas Boreas Hard includes both directions to introduce greater viewpoint variation. For Weather‑KITTI, we use the snow 00, fog 07, and rain 00 as query sets, with their corresponding clear-weather sequences serving as the database. The benchmark includes three leading place recognition models: OverlapTransformer (OT)\cite{ma2022overlaptransformer}, CVTNet\cite{ma2023cvtnet}, and LPSNet~\cite{liu2024lps}, all pretrained on the KITTI dataset and directly evaluated on Boreas and Weather‑KITTI. The denoising models and training configurations are the same as those used in the SS benchmark. 

For the PR evaluation, a match is considered a true recognition if the pose distance is less than 5 meters for the Weather-KITTI, and less than 15 meters for the Boreas dataset. We follow standard metrics widely adopted in prior work~\cite{ma2023cvtnet}, including Recall@1 (R@1), Recall@1\% (R@1\%), and the F1 score. Additionally, we introduce the Mean Robust Score (mRS) to quantify model robustness under adverse weather conditions, defined as:
\begin{equation}
{\text{mRS}} = \frac{1}{N} \sum \frac{\text{M}_{\text{w}}}{\text{M}_{\text{c}}}, \quad \text{M} \in \{\text{R@1}, \text{R@1\%}, \text{F1}\},
\end{equation}
where $\text{M}_{\text{w}}$ and $\text{M}_{\text{c}}$ denote the PR evaluation metrics under adverse and clean weather conditions, respectively. $N$ represents the total number of evaluated metrics across all test sequences. A higher value of mRS reflects greater robustness of the PR model when transitioning from clean to adverse weather conditions

\begin{figure}[!t]
\centerline{\includegraphics[width=\linewidth]{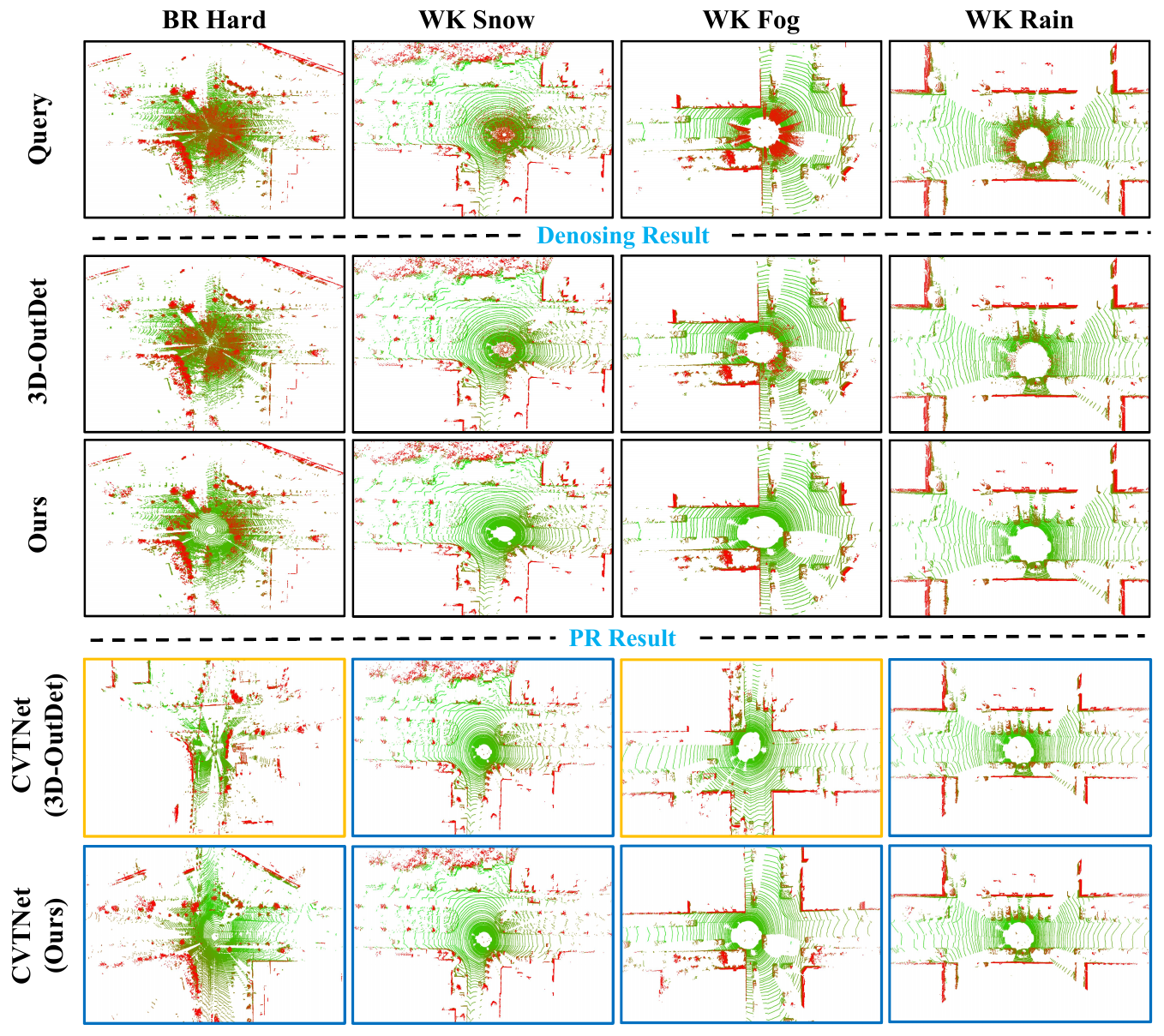}}
	\caption{Qualitative results of the PR benchmark under adverse weather. The PR results show queries processed by different denoising methods and their top‑1 retrieved matches using CVTNet~\cite{ma2023cvtnet}. \textcolor{lightbrown}{Brown boxes} indicate incorrect retrieval, while \textcolor{semanticblue}{blue boxes} denote correct retrievals.}
	\label{LPR_vis}
 \vspace{-0.9em}
\end{figure}

\textbf{Benchmark Results:}
Table~\ref{LPR_BENCH} presents the benchmarking results for the PR task under adverse weather conditions on the Boreas and Weather-KITTI datasets. Across all test sequences, TripleMixer consistently achieves the highest mRS values, demonstrating its effectiveness in enhancing the robustness of PR models under adverse weather conditions. For OT, TripleMixer improves the mRS from the baseline value of 0.15 to 0.31. Similarly, for CVTNet and LPSNet, TripleMixer achieves mRS values of 0.39 and 0.37, respectively, exceeding both the baseline and other denoising models. TripleMixer achieves an average improvement of 56\% across three PR models under adverse weather conditions. Notably, it consistently enhances place recognition performance even in the challenging Boreas Hard scenarios. These results highlight the effectiveness of both TripleMixer and the proposed Weather-KITTI dataset in improving the robustness of downstream PR models against real-world adverse weather conditions. Figure~\ref{LPR_vis} visualizes the result of PR benchmark. Most noise points are concentrated near the LiDAR center. Our TripleMixer achieves effective denoising on both the real‑world Boreas dataset and our Weather‑KITTI dataset, thereby improving PR accuracy.

\begin{table*}[h]
\centering
\footnotesize
\captionsetup{aboveskip=2pt, belowskip=0pt}
\renewcommand\arraystretch{1.1}
\setlength{\tabcolsep}{2.2pt}
\caption{\centering{Benchmarking Results for the OD Task (\%). All OD models are trained on the KITTI dataset~\cite{kitti} and inferred on CADC~\cite{CADC}. Denoising models, trained on Weather-NuScenes, are employed as preprocessing during inference.}}
\begin{tabular}{lcccccccccccccc}
\toprule
\multicolumn{2}{c}{\textbf{Methods}} &\multirow{2}{*}{\textbf{$\text{mAP}$↑}}  & \multicolumn{3}{c}{\textbf{Car (3D)}} & \multicolumn{3}{c}{\textbf{Car (BEV)}} & \multicolumn{3}{c}{\textbf{Pedestrian (3D)}} & \multicolumn{3}{c}{\textbf{Pedestrian (BEV)}} \\

\cmidrule(lr){1-2} \cmidrule(lr){4-6} \cmidrule(lr){7-9} \cmidrule(lr){10-12} \cmidrule(lr){13-15} 

\textbf{OD}  &\textbf{Preprocessing}  & & \textbf{Easy↑} & \textbf{Med↑} & \textbf{Hard↑} 
&\textbf{Easy↑} & \textbf{Med↑} & \textbf{Hard↑}
&\textbf{Easy↑} & \textbf{Med↑} & \textbf{Hard↑}
&\textbf{Easy↑} & \textbf{Med↑} & \textbf{Hard↑}
\\

\midrule
\multirow{4}{*}{{TED-S~\cite{wu2023transformation}}} &Baseline &12.18 &21.85 &20.07 &15.76 &24.16 &21.13 &18.65 &4.26 &4.17 &3.47  &4.54 &4.46 &3.72
\\
&4DenoiseNet~\cite{4DenoiseNet}	&11.91 &22.31 &20.03 &14.88 &25.20 &21.77 &17.51 &4.24 &3.12 &2.81 &4.35 &3.71 &3.01
\\
&3D-OutDet~\cite{3DOutDet}	&12.63 &22.33 &20.05 &16.66 &25.18 &22.74 &18.50 &5.02 &4.12 &3.64 &5.23 &4.31 &3.80
\\
&TripleMixer (Ours)	&\textbf{13.58} &\textbf{24.35} &\textbf{22.06} &\textbf{18.65} &\textbf{27.17} &\textbf{23.73} &\textbf{19.49} &\textbf{5.24} &\textbf{4.20} &\textbf{3.81} &\textbf{5.54} &\textbf{4.73} &\textbf{4.02}
\\

\midrule

\multirow{4}{*}{{PG-RCNN~\cite{koo2023pg}	}} &Baseline &12.53 &22.36 &19.11 &16.67 &24.91 &22.51 &19.89 &5.36 &5.05 &1.99  &5.41 &5.09 &2.04
\\
&4DenoiseNet~\cite{4DenoiseNet}	&12.25 &22.60 &19.13 &16.72 &25.81 &22.30 &19.73 &4.31 &4.17 &1.59 &4.33 &4.21 &1.84
\\
&3D-OutDet~\cite{3DOutDet} &13.43 &23.79 &19.93 &16.11 &26.84 &23.09 &18.96 &6.28 &6.16 &3.80 &6.29 &6.18 &3.84
\\
&TripleMixer (Ours) &\textbf{14.14} &\textbf{24.90} &\textbf{20.23} &\textbf{17.81} &\textbf{28.32} &\textbf{23.43} &\textbf{20.82} &\textbf{6.42} &\textbf{6.39} &\textbf{4.23} &\textbf{6.45} &\textbf{6.43} &\textbf{4.27} 
\\

\midrule 

\multirow{4}{*}{{VoxT-GNN~\cite{zheng2025voxt}	}} &Baseline &10.02 &15.34  &14.24 &11.93 &19.91 &18.78 &15.38
&4.37 &4.16 &3.71  &4.46 &4.25 &3.79 
\\
&4DenoiseNet~\cite{4DenoiseNet}	&9.70 &15.81 &13.84 &11.58 &20.40 &18.23 &14.78 &3.63 &3.48 &3.25 &4.08 &3.70 &3.56
\\
&3D-OutDet~\cite{3DOutDet}	&10.53 &16.59 &14.94 &12.55 &21.18 &19.31 &15.81 &4.62 &4.18 &4.01 &4.85 &4.26 &4.09
\\
&TripleMixer (Ours)	&\textbf{12.33} &\textbf{19.41} &\textbf{16.37} &\textbf{14.13} &\textbf{23.99} &\textbf{20.93} &\textbf{18.09}  &\textbf{6.41} &\textbf{6.38} &\textbf{4.96} &\textbf{6.14} &\textbf{6.12} &\textbf{4.93}
\\

\bottomrule
\label{LOD_BENCH}
\end{tabular}
\vspace{-0.5cm}
\end{table*}

\subsection{ Object Detection (OD) Benchmark} 
\textbf{Benchmark Setting:}
For the OD benchmark, we use the real‑world CADC~\cite{CADC} dataset, which provides object‑level annotations for 3D object detection in a KITTI‑style format, ensuring compatibility with standard evaluation protocols. We evaluate three advanced 3D object detection models, TED‑S~\cite{wu2023transformation}, PG‑RCNN~\cite{koo2023pg}, and VoxT‑GNN~\cite{zheng2025voxt}, using their pretrained versions on the KITTI dataset and directly test them on CADC dataset. Consistent with the SS benchmark, we employ the same denoising models, trained on Weather‑NuScenes and evaluated on CADC, leveraging their shared 32‑beam LiDAR configuration. For evaluation, we follow the KITTI 3D object detection protocol, reporting the mean Average Precision (mAP) over 40 recall positions (R40). We adopt IoU thresholds of 0.7 for cars and 0.5 for pedestrians, while excluding cyclists due to their absence in CADC. Results are reported across three difficulty levels: easy, moderate, and hard.

\begin{figure}[!t]
	\centerline{\includegraphics[width=\linewidth]{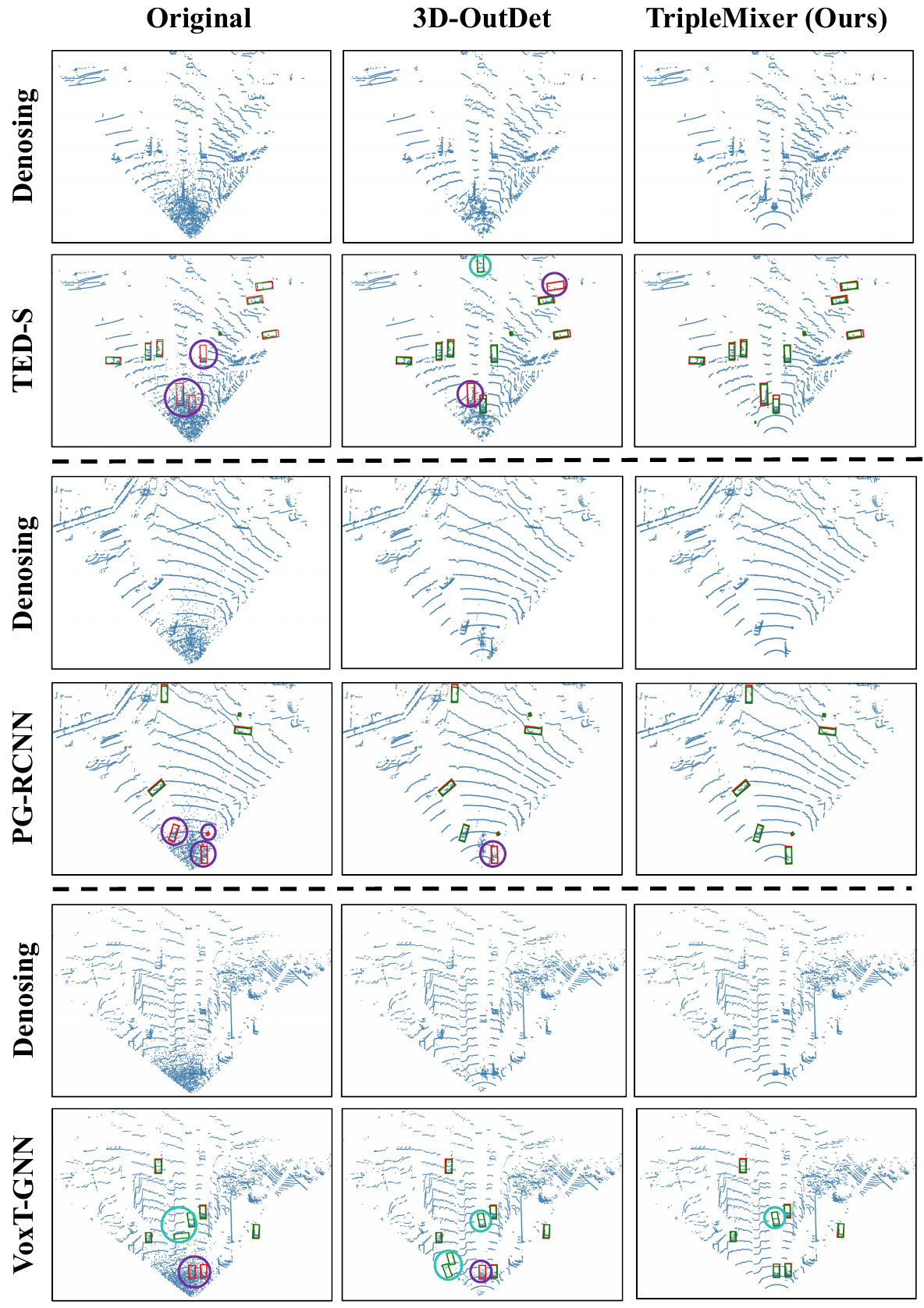}}
	\caption{Qualitative results of the OD benchmark under adverse weather. The “Denoising” rows display outputs from different denoising models, and the subsequent rows present object detection results on the corresponding denoised point clouds.
    \textcolor{red}{Red boxes} indicate ground‑truth objects, \textcolor{green}{green boxes} denote predicted objects. \textcolor{tealframe}{Turquoise  circles} highlight incorrect detections, and \textcolor{customviolet}{pink circles} indicate missed detections.}
	\label{LOD-VIS}
 \vspace{-0.9em}
\end{figure}

\textbf{Benchmark Results:}
The benchmarking results for the OD task under real-world adverse weather conditions on the validation set of CADC~\cite{CADC} dataset, as summarized in Table~\ref{LOD_BENCH}. Evaluations are conducted for both 3D and Bird’s Eye View detection across car and pedestrian categories. The results demonstrate that incorporating TripleMixer as a preprocessing module consistently improves both 3D and BEV detection performance under real-world adverse weather conditions across all OD baselines, including TED-S~\cite{wu2023transformation}, PG-RCNN~\cite{koo2023pg}, and VoxT-GNN~\cite{zheng2025voxt}, with an average performance gain of 16\%. Specifically, when applied to PG-RCNN, it increases the overall mAP from 12.53 to 14.14 and improves 3D detection for the Car category under hard conditions from 16.67 to 17.81. Similarly, with VoxT-GNN, it boosts the mAP from 10.02 to 12.33 and enhances Pedestrian 3D detection under hard conditions from 3.71 to 4.96. In contrast, 4DenoiseNet fails to improve detection performance, similar to its results in the SS task. Its limited denoising ability hinders performance on detail-sensitive tasks. These results show that TripleMixer benefits from our Weather-NuScenes dataset, enabling more effective spatial restoration and consistently improving OD performance under real-world extreme weather across different backbones and categories. Figure~\ref{LOD-VIS} illustrates the performance of OD models with different denoising preprocessing under adverse weather. Models without denoising exhibit numerous false positives and missed detections. Compared with 3D‑OutDet, our TripleMixer produces cleaner point clouds, enabling OD models to achieve higher detection accuracy and reducing both false positives and missed detections.



\begin{table}[ht]
\captionsetup{justification=centering, singlelinecheck=false}
\centering
\caption{{Ablation Results for Each Layer of TripleMixer.}}
\setlength{\tabcolsep}{3pt}
\begin{tabular}{cccccccc}
\toprule
\multirow{2}{*}{\textbf{Experiment}} & \multirow{2}{*}{\textbf{GMX}} & \multirow{2}{*}{\textbf{FMX}} & \multirow{2}{*}{\textbf{CMX}} & \multicolumn{4}{c}{\textbf{Metric(\%)}} \\
\cmidrule(lr){5-8}
 &  & & & \textbf{Precision} & \textbf{Recall} & \textbf{F1} & \textbf{mIOU} \\
\midrule
\uppercase\expandafter{\romannumeral1} & &\checkmark &\checkmark &95.44 &93.86 &94.64  &89.83    \\
\uppercase\expandafter{\romannumeral2} &\checkmark  & &\checkmark   &95.11 &93.65  &94.37  &89.35  \\
\uppercase\expandafter{\romannumeral3} & \checkmark &\checkmark &   &96.34  &93.71 &95.01  &90.48 \\
\uppercase\expandafter{\romannumeral4} & \checkmark &\checkmark &\checkmark &\textbf{96.38}  &\textbf{93.93}  &\textbf{95.13} &\textbf{90.73}  \\
\bottomrule
\end{tabular}
\label{ablation_component}
\end{table}

\begin{table}[ht]
\captionsetup{justification=centering, singlelinecheck=false}
\centering
\caption{{Ablation Results for XY Projection Resolutions with Z Resolution Fixed at 32.}}
\begin{tabular}{cccccc}
\toprule
\textbf{XY-Resolution} & \textbf{Precision↑} & \textbf{Recall↑} & \textbf{F1↑} & \textbf{mIoU↑}   \\
\midrule
{[64,64]} &96.24  &93.63  &94.91  &90.32   \\
{[128,128]} &\textbf{96.55}  &93.57  & 95.03 &90.54  \\
{[256,256]} & 96.38 & \textbf{93.93} & \textbf{95.13} & \textbf{90.73}   \\
{[384,384]} &96.33  &98.83  &95.06  &90.59   \\
{[512,512]} &96.25  &98.85  &95.03  &90.53   \\
\bottomrule
\label{plane_reso_xy}
\end{tabular}
\vspace{-1.4em}
\end{table}


\begin{table}[ht]
\captionsetup{justification=centering, singlelinecheck=false}
\centering
\caption{{Ablation Results for Z Projection Resolutions with XY Resolutions Fixed at [256, 256]}}
\begin{tabular}{cccccc}
\toprule
\textbf{Z-Resolution} & \textbf{Precision↑} & \textbf{Recall↑} & \textbf{F1↑} & \textbf{mIoU↑}   \\
\midrule
{[16]} &96.43  &93.73  &95.06  &90.57  \\
{[24]} &\textbf{96.55}  &93.70  &95.10  &90.66   \\
{[32]} & 96.38 & \textbf{93.93} & \textbf{95.13} & \textbf{90.73}   \\
{[40]} &96.43  &93.69  &95.04  &90.54   \\
{[48]} &96.46  &93.59  &95.00  &90.48    \\
\bottomrule
\label{plane_reso_z}
\end{tabular}
\vspace{-1.4em}
\end{table}

\begin{table}[ht]
\captionsetup{justification=centering, singlelinecheck=false}
\centering
\caption{{Ablation Results for Wavelet Decomposition Levels in FMX Layer}}
\begin{tabular}{ccccccc}
\toprule
\textbf{Methods} & \textbf{Precision↑} & \textbf{Recall↑} & \textbf{F1↑} & \textbf{mIoU↑}  \\
\midrule
{w/o Wavelet} &95.11  &93.65  &94.37  &89.35  \\
{w/ Wavelet-1} & 96.33 &93.54  &94.91  &90.32  \\
{w/ Wavelet-2} & \textbf{96.38} & \textbf{93.93} & \textbf{95.13} & \textbf{90.73}  \\
{w/ Wavelet-3} &96.16  &93.80  &94.96  &90.41   \\
\bottomrule
\label{wavelet}
\end{tabular}
\vspace{-1.2em}
\end{table}

\begin{table}[ht]
\centering
\caption{\centering{Ablation Results of Loss Functions.}}
\setlength{\tabcolsep}{5pt}
\begin{tabular}{ccccccc}
\toprule
\multirow{2}{*}{\textbf{Experiment}} & \multirow{2}{*}{\textbf{$\boldsymbol{\mathcal{L}_{ce}}$}} & \multirow{2}{*}{\textbf{$\boldsymbol{\mathcal{L}_{{lovasz}}}$}} & \multirow{2}{*}{\textbf{$\boldsymbol{\mathcal{L}_{{wr}}}$}} & \multicolumn{2}{c}{\textbf{Metric(\%)}} \\
\cmidrule(lr){5-6}
 &  & & & \textbf{F1} & \textbf{mIOU} \\
\midrule
\uppercase\expandafter{\romannumeral1} &\checkmark  &  &   &94.84  &90.19  \\
\uppercase\expandafter{\romannumeral2} & \checkmark &\checkmark &   &95.05  &90.56  \\
\uppercase\expandafter{\romannumeral3} & \checkmark & &\checkmark   &94.98  &90.45 \\
\uppercase\expandafter{\romannumeral4} & \checkmark &\checkmark &\checkmark &\textbf{95.13}  &\textbf{90.73}   \\
\bottomrule
\end{tabular}
\label{loss}
\end{table}

\section{Discussion}

\subsection{Ablation Study} 
This section presents ablation studies to evaluate the impact of key components in the TripleMixer model, including three primary layers, projection plane resolution, wavelet decomposition levels and loss functions. All ablation experiments are conducted on the real-world WADS\cite{dsor} dataset, with same experimental settings as in Section \uppercase\expandafter{\romannumeral5}.

\textbf{Effects of Each Layer.} To testify the effectiveness of each Layer of TripleMixer, we conduct an extensive ablation study and list the result in Table \ref{ablation_component}. Experiments \uppercase\expandafter{\romannumeral1}-\uppercase\expandafter{\romannumeral3} represent the accuracy of our TripleMixer model after sequentially removing the GMX Layer, the FMX Layer, and the CMX Layer, respectively. From Experiments \uppercase\expandafter{\romannumeral1} and \uppercase\expandafter{\romannumeral4}, we observe that the GMX Layer contributes significantly, resulting in F1 and mIoU scores of 0.49 and 0.90, respectively. Similarly, Experiments \uppercase\expandafter{\romannumeral2} and \uppercase\expandafter{\romannumeral4} indicate that the FMX Layer enhances performance, yielding F1 and mIoU improvements of 0.65 and 1.19, respectively. Finally, the impact of the CMX Layer is evident from Experiments \uppercase\expandafter{\romannumeral3} and \uppercase\expandafter{\romannumeral4}, showing F1 and mIoU increases of 0.12 and 0.25, respectively. These results demonstrate that the GMX Layer, the FMX Layer, and the CMX Layer are all crucial for improving model performance, each contributing significantly to the overall effectiveness of our TripleMixer model.

\textbf{Effects of Projection Plane Resolution.} In the FMX layer, 3D geometric features are projected onto three 2D planes along the X, Y, and Z axes. Adaptive wavelet decomposition is then applied to extract multi-scale features from each plane. The initial resolution of these projections plays a key role in determining the quality of the extracted features. To evaluate its impact, we conduct ablation studies on different projection resolutions. Considering the asymmetric LiDAR scanning ranges, we separately investigate the XY and Z directions. First, we fix the Z-direction resolution at 32 and explore various resolutions in the XY plane. As shown in Table~\ref{plane_reso_xy}, the highest mIoU is achieved at [256, 256], while both lower ([64, 64]) and higher ([512, 512]) resolutions lead to performance drops. Based on this, we fix the XY resolution to [256, 256] and examine different Z-direction resolutions. Table~\ref{plane_reso_z} shows that a resolution of 32 again yields the best performance, with other values resulting in suboptimal accuracy. These findings confirm that the projection resolution of [256, 256, 32] provides the best balance between feature granularity and computational efficiency, and is adopted in all practical experiments.

\textbf{Effects of Wavelet Decomposition Levels.} We evaluate the impact of different wavelet decomposition levels in the Frequency Mixer (FMX) on model performance. With the projection resolution fixed at [256, 256, 32] along the XYZ axes, we set the maximum wavelet decomposition level to 3, as higher levels reduce the Z-direction resolution to less than or equal to 2, limiting the extraction of meaningful features. We examine decomposition levels ranging from 0 to 3. As shown in Table~\ref{wavelet}, omitting wavelet decomposition yields the lowest accuracy, underscoring its importance in feature extraction. While moderate decomposition improves performance, excessive decomposition introduces information loss. The best results are obtained when the wavelet decomposition level is set to 2, confirming its effectiveness in balancing detail preservation and multi-scale representation.

\textbf{Effects of Loss Functions.} Our model's loss function primarily consists of three components: the cross-entropy loss, the lovasz loss, and the wavelet regularization terms. We use the cross-entropy loss as the baseline to test the impact of different loss functions on model accuracy. As shown in Table \ref{loss}, after adding only the lovasz loss, the mIoU and F1 scores of our TripleMixer model improved by 0.37 and 0.21, respectively. Additionally, after incorporating only the wavelet regularization terms, the model's mIoU and F1 scores increased by 0.26 and 0.14, respectively. When using the combination of all three loss functions, our denoising model achieved the highest mIoU and F1 scores.

\begin{table}[t] 
\captionsetup{justification=centering, singlelinecheck=false}
\centering
\caption{Comparison of Model Complexity, Runtime, and Peak Memory Usage.}
\setlength{\tabcolsep}{4pt}
\label{runtime_memory}
\begin{tabular}{lccc}
\toprule
\textbf{Method} & \textbf{Parameters (M)} & \textbf{Runtime (ms)} & \textbf{Memory (GB)} \\
\midrule
Cylinder3D~\cite{cylinder3d} &49.52  &127 &0.92 \\
LSK3DNet\cite{feng2024lsk3dnet} &28.84  &89 &0.78 \\
DSOR~\cite{dsor} &-- &253  &--  \\
DROR~\cite{dror} &--  &199  &-- \\
4DenoiseNet\cite{4DenoiseNet} &0.57 & 96 & 2.47 \\
3D-OutDet\cite{3DOutDet} & \textbf{0.01} & 82 &0.33  \\
\textbf{TripleMixer (Ours)} & 1.02 & \textbf{61} &\textbf{0.21}  \\
\bottomrule
\end{tabular}
\end{table}

\subsection{Efficiency Comparisons} Table~\ref{runtime_memory} compares the model complexity, runtime, and peak memory usage during inference across different methods. General point cloud segmentation models adapted for denoising, such as Cylinder3D~\cite{cylinder3d} and LSK3DNet~\cite{feng2024lsk3dnet}, have substantially higher parameter counts of 49.52M and 28.84M and memory footprints of 0.92GB and 0.78GB. Traditional statistical denoising methods DSOR~\cite{dsor} and DROR~\cite{dror} are computationally expensive with long runtimes. 3D‑OutDet\cite{3DOutDet} has the smallest parameter count of 0.01M but a relatively high runtime due to additional k‑NN preprocessing. 4DenoiseNet~\cite{4DenoiseNet} shows the highest peak memory usage of 2.47GB as it aggregates spatiotemporal features from multiple frames. In contrast, our TripleMixer achieves a favorable balance with a runtime of 61ms and the lowest peak memory usage of 0.21GB, demonstrating high efficiency.

\section{Conclusion}
In this work, we proposed TripleMixer, a robust plug‑and‑play denoising network that can be seamlessly integrated into LiDAR perception pipelines. We also introduced two large‑scale simulated LiDAR datasets, Weather‑KITTI and Weather‑NuScenes, encompassing diverse adverse weather conditions. Building on these contributions, we established four benchmarks: Denoising, Semantic Segmentation (SS), Place Recognition (PR), and Object Detection (OD), providing a unified framework for evaluating denoising models and their impact on downstream perception tasks. Our model was trained exclusively on the proposed datasets, and all perception models were directly evaluated on real‑world adverse‑weather datasets without retraining. Extensive experiments demonstrated that TripleMixer achieves state‑of‑the‑art denoising performance and substantially improves SS, PR, and OD accuracy under real‑world adverse weather, underscoring the strong generalization capability of both the proposed model and datasets.

\textbf{Future Work.} While the multi‑branch mixer design of TripleMixer effectively suppresses noise, its relatively high parameter count may hinder deployment in resource‑constrained settings. Future work will focus on developing a more lightweight denoising network. Furthermore, while synthetic weather data provides valuable scalability and controllability, it may not fully capture the subtle complexities of real‑world weather conditions. We will explore more realistic simulation techniques and expand real‑world adverse‑weather datasets to further improve model generalization.



\bibliographystyle{IEEEtran}

\bibliography{IEEEreference}

\end{document}